\documentclass[journal,comsoc]{IEEEtran}
\usepackage[T1]{fontenc}
\usepackage[T1]{fontenc}

\usepackage[latin5]{inputenc}
\usepackage{times}
\usepackage[figuresright]{rotating}
\usepackage{float}
\usepackage{subcaption}
\usepackage{url}
\usepackage{setspace}
\usepackage{balance}

\usepackage{multicol}
\usepackage{multirow}
\usepackage{color}
\usepackage{graphicx}
\usepackage[implicit=false]{hyperref}
\usepackage{epsfig}
\usepackage{amsmath, amssymb}
\usepackage{comment}

\usepackage{capt-of}
\usepackage{multirow}
\usepackage{array}
\usepackage{caption} 
\usepackage{pslatex}
\usepackage{enumitem}
\usepackage{cite}
\usepackage{tcolorbox}
\usepackage{tabularx}
\tcbset{tab2/.style={colback=red!5!white,colframe=red!50!black,colbacktitle=red!40!white,
coltitle=black,center title}}

\usepackage{colortbl}

\makeatletter
\newcommand{\thickhline}{%
    \noalign {\ifnum 0=`}\fi \hrule height 1.5pt
    \futurelet \reserved@a \@xhline
}
\newcolumntype{"}{@{\hskip\tabcolsep\vrule width 1pt\hskip\tabcolsep}}
\makeatother
\newcolumntype{?}{!{\vrule width 1.5pt}}
\ifCLASSINFOpdf
\else
\fi
\usepackage{amsmath}
\usepackage[cmintegrals]{newtxmath}
\usepackage{graphicx}
\usepackage{acronym}

\begin{document}
%
\title{A Comprehensive Survey on Radio Frequency (RF) Fingerprinting: Traditional Approaches, Deep Learning, and Open Challenges}
%
%
%

\author{
        Anu~Jagannath,
        Jithin~Jagannath,
        and~Prem Sagar Pattanshetty Vasanth Kumar
\thanks{Anu Jagannath is with Northeastern University and ANDRO's Marconi-Rosenblatt AI/ML Innovation Laboratory.}
\thanks{Jithin Jagannath is with University at Buffalo and ANDRO's Marconi-Rosenblatt AI/ML Innovation Laboratory..}
\thanks{Prem Sagar Pattanshetty Vasanth Kumar is with ANDRO's Marconi-Rosenblatt AI/ML Innovation Laboratory.}}

\maketitle

\begin{abstract}
Fifth generation (5G) network and beyond envision massive Internet of Things (IoT) rollout to support disruptive applications such as extended reality (XR), augmented/virtual reality (AR/VR), industrial automation, autonomous driving, and smart everything which brings together massive and diverse IoT devices occupying the radio frequency (RF) spectrum. Along with the spectrum crunch and throughput challenges, such a massive scale of wireless devices exposes unprecedented threat surfaces. RF fingerprinting is heralded as a candidate technology that can be combined with cryptographic and zero-trust security measures to ensure data privacy, confidentiality, and integrity in wireless networks. Motivated by the relevance of this subject in the future communication networks, in this work, we present a comprehensive survey of RF fingerprinting approaches ranging from a traditional view to the most recent deep learning (DL)-based algorithms. Existing surveys have mostly focused on a constrained presentation of the wireless fingerprinting approaches, however, many aspects remain untold. In this work, however, we mitigate this by addressing every aspect - background on signal intelligence (SIGINT), applications, relevant DL algorithms, systematic literature review of RF fingerprinting techniques spanning the past two decades, discussion on datasets, and potential research avenues - necessary to elucidate this topic to the reader in an encyclopedic manner.
\end{abstract}


%
\IEEEpeerreviewmaketitle

\section{Introduction}

\IEEEPARstart{R}{adio} frequency (RF) fingerprinting - a form of signal intelligence - refers to the methodology whereby the hardware intrinsic characteristics of the transmitter which are unintentionally embedded in the transmitted waveform are extracted to aid the identification of the transmitter hardware by a passive receiver. Due to its unique ability to identify transmitting device, RF fingerprinting is envisioned as a key enabler for device authentication and access control to reduce the vulnerability of beyond 5G wireless networks to node forgery or insider attacks \cite{device_fingerprint}. 

With the proliferation of wireless devices and the increased adoption of Internet-of-Things (IoT) devices for smart home, industrial automation, smart metering, etc., the beyond 5G network is expected to support ultra-dense device connectivity which is $10\times$ that of 5G \cite{Ajagannath6G2020}. Moreover, with such overwhelming device density the threat surfaces of the network are bound to increase. Therefore, security and privacy are the crucial inevitable aspects beyond 5G (6G) will need to address. Especially the 6G enabling technologies such as ultra-massive multiple-input multiple-output (UM-MIMO), visible light communication (VLC), terahertz (THz) communication, among others, introduce new realm of security challenges. Even in 5G networks, the OpenFlow implementation of the software defined network (SDN) makes it vulnerable to attacks from malicious applications. Further, the network function virtualization (NFV) presents security risks as the function is being migrated from one platform to another \cite{WANG2020281}. With the envisioned device density of the beyond 5G network, such vulnerabilities will only increase. The security threats can perhaps be best attributed to two causes; massive device density and diversity with respect to the applications as well as the hardware.

The hardware intrinsic features of device form the fingerprint or the signature unique to that device. RF fingerprinting is consequently viewed as the prospective enablers to address and mitigate the access control and device authentication challenges of the beyond 5G network. For the purpose of clarity, we define RF fingerprinting as a composite of three steps; feature identification, feature extraction, and device identification. It must be emphasized that these features are location-independent and ingrained to the chipset. Specifically, the imperfections in manufacturing the microcircuit parts such as power amplifiers, filters, clocks, etc., lead to broad variations in the phase offset, clock skew, among others. Another aspect that could serve as features are the vendor-specific implementations of wireless standards \cite{device_fingerprint}. But such features could easily vary with firmware and software upgrades of the chipset. Clearly, the device-specific features would serve as a pronounced invariant feature set. 

Despite several device fingerprinting works, a comprehensive survey encompassing the evolution of fingerprinting algorithms from principled to deep learning based approaches is lacking. The contributions and scope of this article is discussed in detail here to portray its relevance in the present era of evolving wireless networks.

\begin{figure*}[h]
\centering
\includegraphics[width=1.99\columnwidth]{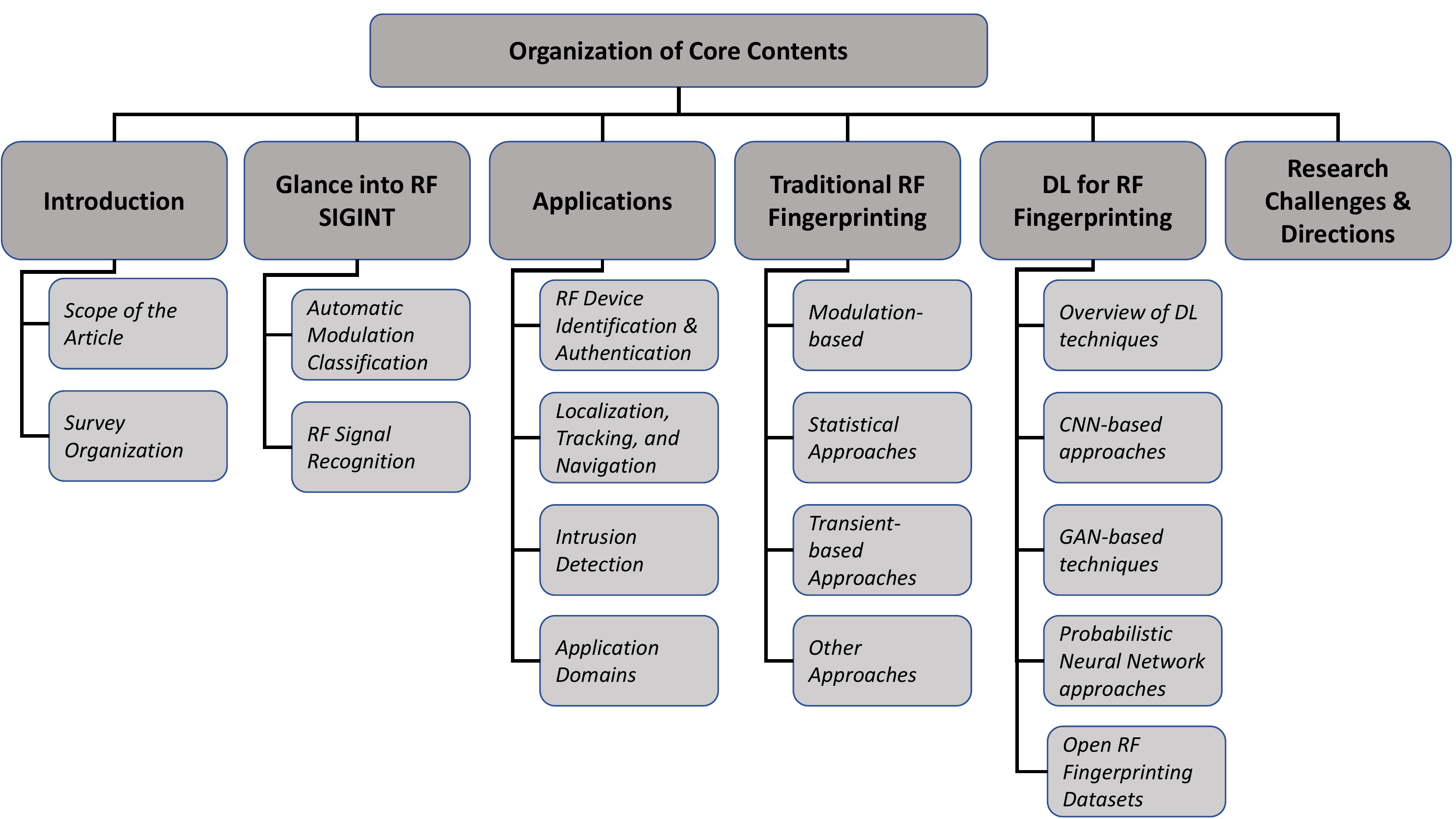} 
\caption{Overview of the organization of the article}
\label{fig:Organization}
\end{figure*}

\subsection{Scope of the Article}
The objective of this article is to present a comprehensive view of the state-of-the-art wireless device fingerprinting algorithms while also provide sufficient background on the subsidiary signal intelligence domains - modulation and wireless protocol classification. 
Although there has been numerous articles on deep learning for other RF signal intelligence approaches (modulation and wireless protocol classification) \cite{XuLiWang2019,ICAMCNet,ANN3,selim2017spectrum_monitoring,Jagannath19MLBook,wirelessInterference,wirelesstech,OSheaCNN2,JagannathAdHoc2019,JagannathMTL,Robinson2020ADCC}, a comprehensive presentation spanning conventional principled approaches as well as supervised deep learning for RF fingerprinting is lacking. We attempt to bridge this gap by discussing the following key aspects:
\begin{enumerate}
    \item A succinct and categorized layout of the related research in the field of RF signal intelligence (SIGINT) to provide relevant background to the reader. Here, we present a preview of the various methods - spanning conventional and deep learning - for automatic modulation classification (AMC), wireless protocol recognition, before going to a illustrative discussion on the RF fingerprinting applications as well as approaches.
    \item Some of the key application domains of RF signal intelligence in this emerging revolutionary communication era where billions of wireless devices including diverse IoT emitters coexist. This aspect presents the viewer with critical wireless network application areas for a practical insight of the presented RF signal intelligence methods. 
    \item A qualitative discussion on the traditional approaches for RF fingerprinting categorized into modulation, statistical, transient, wavelet, and other approaches.
    \item A deep dive into the state-of-the-art deep learning methods for RF fingerprinting including an overview of the prominent deep learning approaches in order to assist researchers in applying them for RF signal intelligence.
    \item A detailed account of the various open-source datasets tailored to equip researchers with comprehensive knowledge to delve into applied RF fingerprinting research.
    \item We motivate further research in this realm by presenting open research challenges and future directions.
\end{enumerate}

We emphasize here that unlike existing surveys, our article is comprehensive in presenting all aspects of RF fingerprinting comprising a glance into background on RF signal intelligence, the evolution towards deep learning approaches for RF fingerprinting including the progress on conventional principled methods. For completeness and to benefit beginners, we provide a tutorial of the relevant deep learning techniques.

Most of the existing surveys related to RF fingerprinting presents only a narrow scope. Specifically, a qualitative analysis of all RF signal intelligence aspects including AMC, wireless protocol recognition, and a quantitative discussion on key deep learning approaches have not been widely investigated to date in the current literature. In \cite{baldini_phyFingerprint}, the various techniques of identifying a mobile phone by fingerprinting the built-in components, such as camera, micro-electro-mechanical systems, speakers, microphone, and RF frontends have been discussed. In contrast, we attempt to cover all RF emitters such as software-defined radios (SDRs), unmanned aerial vehicles (UAVs), and other consumer-off-the-shelf (COTS) devices including mobile phones. The survey in \cite{guo_survey} presents a short account of the RF fingerprint extraction and authentication methods with an emphasis on device authenticity - legal or illicit. Another survey in \cite{gnss} reviews spoofer detection methods that leverages RF fingerprinting with special emphasis on Global Navigation Satellite System (GNSS) emitters. Although this work presents a broader scope in contrast to \cite{baldini_phyFingerprint} and \cite{guo_survey}, it lacks a thorough presentation of all aspects of RF signal intelligence. One other survey in \cite{rffsurvey} discusses the taxonomy of wireless device fingerprinting along with brief account on fingerprinting algorithms that are classical white-list and unsupervised learning-based. Our article goes beyond these works in providing the reader all aspects of RF signal intelligence including crossovers between traditional and deep learning based RF signal intelligence approaches. Unlike existing surveys which only provides a very brief (2-3 sentences) discussion on the reviewed articles, we dive into the reviewed works in the vast literature to provide a succinct excerpt on each. Further, the application scope of these signal intelligence techniques are not elaborated in any other existing surveys. 
The ultimate objective of this article is to provide an encyclopedic guide of RF fingerprinting that encompasses the basics of key supervised deep learning techniques as well as an extensive review of the state-of-the-art RF signal intelligence. 

\subsection{Survey Organization} 

We structure our article in an organized hierarchical manner: Section \ref{sec:glance} introduces the readers to the two subsidiary domains of RF signal intelligence and reviews the traditional as well as deep learning based automatic modulation and wireless protocol classification. 
The key application areas of the discussed RF fingerprinting methods are briefly discussed in section \ref{sec:apps} to supplement practical insight to the researchers and practitioners allowing them to explore the applicability. We begin the RF fingerprinting survey by elaborating on the principled approaches first in section \ref{sec:traditional}. We have categorized the traditional approaches based on the fingerprinted characteristics into modulation, statistical, transient, wavelet, and other miscellaneous methods to enable a sectioned and comparative discussion of the vast literature on traditional techniques. Next, we present an illustrative discussion on the state-of-the-art deep learning-based RF fingerprinting techniques in section \ref{sec:dlrff}. We have segmented this section into two where the first part reviews the key deep learning concepts to present contextual walk-through for the readers, followed by the second part which shows how these deep learning techniques are applied to the RF fingerprinting domain. Further, we educate the readers on the available open-source datasets for training deep learning models to perform RF fingerprinting. Finally, we aim to spur future research in this domain by summarizing a few open questions and challenges in section \ref{sec:open}. We also layout the organization in a pictorial manner in Figure \ref{fig:Organization}. 

\section{Glance into RF Signal Intelligence}
\label{sec:glance}

RF signal intelligence is defined as the field of research and application that focuses on extracting signal characteristics such as modulation, bandwidth, center frequency, protocols, emitter identity, among others from unknown RF signals in the spectrum of interest. This extraction can be performed under various levels of cooperation or prior knowledge based on the application at hand. The most challenging version is under the assumption of no prior information or cooperation which is often referred to as blind RF signal intelligence. 

This area of research is further divided into different categories based on the task performed. Perhaps the most popular and widely researched task is that of AMC and then wireless signal/protocol classification. One common theme between these classification tasks are the fact that the signals itself are evidently different from each other in these classification tasks making them a relatively easier task compared to RF fingerprinting where identical devices could be transmitting same waveform with identical configurations. Here for the benefit of the readers, we provide the background information of the two most common signal intelligence classes (AMC and signal type classification). This is also for the readers to relate to the overall RF signal intelligence research domain while reviewing the more in-depth survey of RF fingerprinting approaches. 

\subsection{Automatic modulation classification}

\begin{figure}[h]
\centering
\includegraphics[width=0.99\columnwidth]{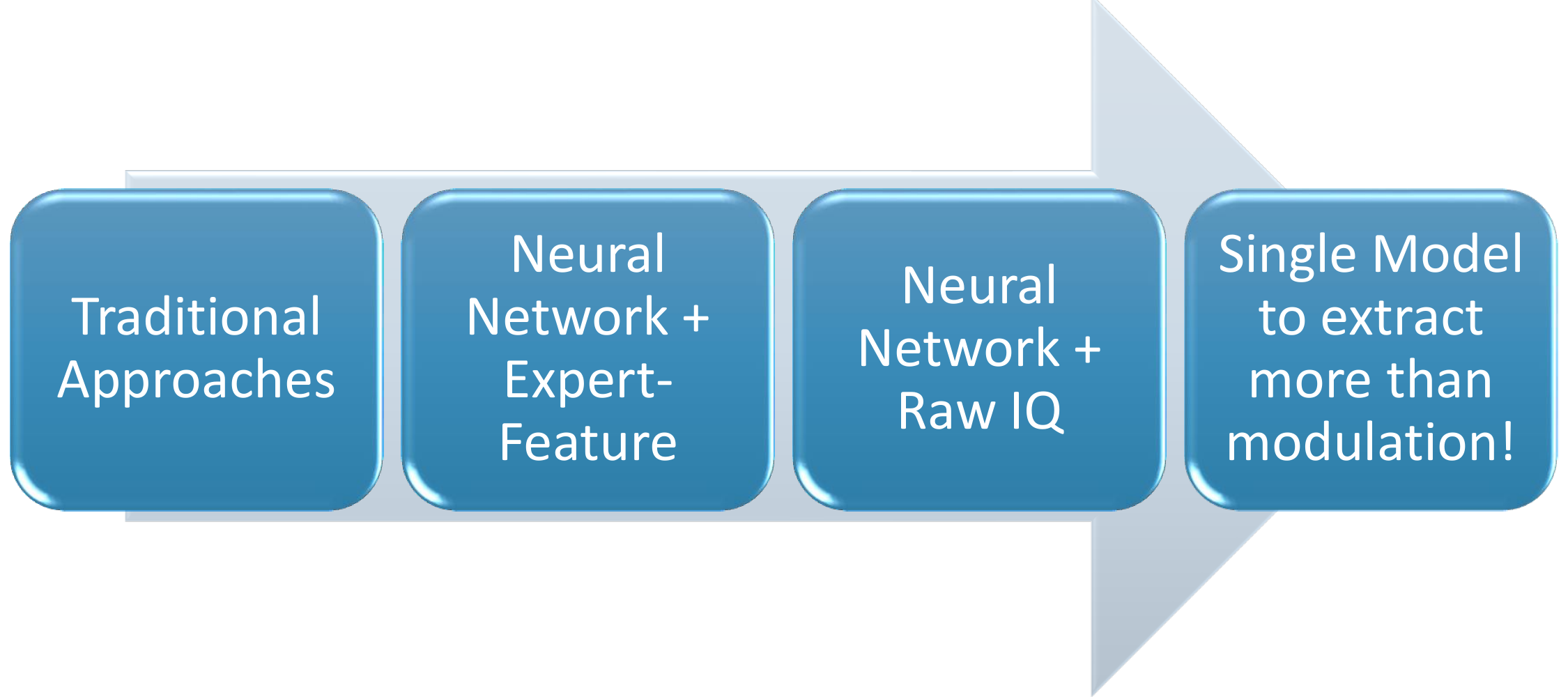} 
\caption{Evolution of AMC Approaches}
\label{fig:Evolution}
\end{figure}

As discussed earlier, due to the extensive attention this area of research has garnered, we organize this section based on the evolution seen in AMC techniques depicted in Figure \ref{fig:Evolution}.

\textbf{Traditional Approaches:} AMC can be broadly categorized into two classes; (i) likelihood-based methods \cite{ML_Hameed, LB_Jianping, Dis_AMC, ML_Su_2,Varshney_1,Varshney_2} and (ii) feature-based \cite{Feature_1, Cumulant_Chang, FB_Sudhan_2017, Han_1}. There have been several attempts to combine the two approaches to possibly extract the benefits of both approaches \cite{Jagannath17CCWC}. Likelihood-based approaches can provide optimal performance in the Bayesian sense but are often computationally demanding \cite{Jagannath17CCWC, Foulke}. On the other hand, feature-based classifiers can provide near optimal performance while being computationally efficient if carefully designed. Note here that the requirement of being ``carefully designed" is perhaps the weakness of traditional feature-based approaches. It is often possible to design the classifier which performs extremely well under certain assumptions in simulations or laboratory settings but fail under real-world scenarios or when the operational environment changes. In other words, for AMC to be suitable for real-world approaches, it is important for the classifiers to generalize well to various operating scenarios and environments. 

\textbf{Neural network with expert-feature:} Since the problem structure of feature-based classifiers are similar to the function approximation schema of the recently revitalized supervised machine learning, it was inevitable for these techniques to be leveraged for AMC. Consequently, in recent years, different machine learning techniques have been employed to determine the modulation format of the unknown signal via classification. During the initial stages of applying supervised learning for AMC, feature-engineered methodology was adopted as opposed to utilizing raw in-phase and quadrature (IQ) samples. 
This includes the use of support vector machines (SVMs) \cite{SVM} and ANNs \cite{ANN2,ANN,ANN3}. In \cite{ANN2}, the authors perform a twelve-class modulation classification with high accuracy over a wide range of signal-to-noise ratio (SNR) values using a multilayer perceptron (MLP). In \cite{ANN}, the authors evaluate two different ANN architectures trained by the backpropagation method using the standard gradient descent (GD) learning algorithm by using six features. Similarly, \cite{ANN3} achieves high accuracy under low SNR conditions in identifying eight modulation schemes. All these studies are limited to simulations and not evaluated on actual hardware. In \cite{Foulke}, authors elaborate the confronted challenges while transitioning their solution from simulation to hardware implementation. In short, due to the assumptions and unanticipated signal distortions that are overlooked during simulations, over-the-air performance of AMC techniques may experience degradation in real deployment. 

The superior feature extraction capability of convolutional neural networks (CNNs) in contrast to ANNs led to several works leveraging CNNs for modulation or signal classification \cite{AlexGoogleNet_AMC, vhf_amc,ICAMCNet,oshea2018,Ajagannath2022SPIE, wirelessInterference,wirelesstech}. The authors of \cite{AlexGoogleNet_AMC} evaluated the performance of CNNs - GoogLeNet \cite{googlenet} and AlexNet \cite{alexnet} architectures - in predicting modulation formats on a dataset comprising eight classes by feeding constellation images as input. However, the models demonstrated sensitivity to image preprocessing factors such as image resolution, cropping size, selected area, etc., and achieved an accuracy below 80\% at 0 dB SNR. We profess and attribute that this could be due to the adoption of heavy architectures suited for computer vision problems rather than for the RF application. A feed-forward feature-based neural network \cite{Jagannath18ICC} was shown to achieve a classification accuracy of 98\% on seven modulations on a USRP software-defined radio testbed. Time-frequency images were used as input for a CNN architecture to identify seven radar waveform classes in \cite{radar_recog}. Along a similar trend, cyclic spectrum images were used as CNN input to obtain a seven-class modulation recognition accuracy of 95\% at SNRs above 2 dB in \cite{vhf_amc}. We would like to emphasize here that these works rely on handcrafted features to train the neural network which limits the generalization capability of the network as it could have from raw IQ samples.

\textbf{Neural network with raw IQ:} A CNN architecture which classifies $5$ communication waveforms by utilizing raw IQ samples was explored in \cite{XuLiWang2019}. Although the model achieves a $100\%$ accuracy it considers very limited number of waveforms of the same carrier frequency and bandwidth. The authors of \cite{ICAMCNet} trained a CNN to achieve an accuracy of 83.4\% at 18 dB in classifying 11 modulations by feeding raw IQ samples. In \cite{oshea2018}, a modified ResNet architecture was shown to achieve a 95.6\% accuracy at 10 dB by learning from raw IQ samples in identifying 24 modulation classes.

\subsection{RF signal recognition}

Wireless signal recognition is a signal (wireless standard or protocol) recognition method which involves identifying the wireless standard with which the RF waveform is generated. The authors of  \cite{wirelessInterference} studied wireless interference detection by performing a 15-class identification comprising three wireless standards - IEEE 802.11 b/g, IEEE 802.15.4, and IEEE 802.15.1 - occupying different frequency channels. In a similar sense, \cite{wirelesstech} adopted a CNN architecture to address the spectrum crunch in the industrial, scientific, and medical (ISM) band by classifying seven classes belonging to Zigbee, WiFi, Bluetooth, and their cross-interferences. However, the model required operation in a high SNR regime for a 93\% accuracy. In \cite{LSSUAV2017shi}, the authors use a distance-based support vector data description (SVDD) algorithm to recognize low, slow, and small unmanned aerial vehicles (LSSUAVs) among the signals in the 2.4 GHz band by generating a hash fingerprint. The proposed method recognized LSSUAV signals without any mistakes and falsely recognized IEEE 802.11b and IEEE802.11n signals as LSSUAV $13.5\%$ and $0\%$ of the time respectively in an indoor environment. Similarly, the authors \cite{zuo2021Recognition_UAV} investigate recognition of UAV video signals in the presence of WiFi interference. Using random forest classifier, the authors show that the method can recognize UAV video signal in presence of WiFi interference with an accuracy of $100\%$ indoors and $96.26\%$ when the UAV is 2 km from the receiver. In \cite{selim2017spectrum_monitoring}, the authors implement a CNN model to identify the presence of radar signals in the radio spectrum with interference from LTE and WLAN signals. The authors achieve a classification accuracy of $99.6\%$ while using amplitude and phase shift components of the signals in the dataset. The authors in \cite{kulin2018end_to_end} train CNN classifiers using time domain features to recognize WiFi, Zigbee, and Bluetooth devices operating in the 2.4 GHz band. The results demonstrate that the proposed method is capable of recognizing with an accuracy $\geq95\%$ for SNR greater then 5 dB.

\subsection{Single model to extract more than modulation}

A multi-task learning (MTL) model that can learn to recognize more than one task - modulation and signal (protocol) recognition - was proposed for the first time in \cite{AJagannath21ICC,AJagannath22PHYCOM}. This was the first work to consider both radar and communication waveforms to address the diverse and heterogeneous signal types encountered in practical deployment. Here, the authors train a CNN to perform two related tasks based on a single raw IQ input. The two tasks are assigned weights to formulate the weighted sum loss function and the model was trained with backpropagation. The authors emphasize the significance of designing lightweight models from the inception and provide real-world experimental evaluation with over-the-air collected waveforms under varying signal strengths. The evaluations demonstrated high-speed inferences The lightweight MTL performs faster inferences at the rate of 8.4 ms on an Intel Core i5-3230M CPU, consuming up to 90.5\% lesser memory requirement in contrast to the benchmark. Further, the model was further compressed by performing INT8 quantization to showcase the computational savings for resource-constrained edge deployment platforms. The uncompressed 32-bit floating point (FP32) model was compressed 11.8$\times$ by INT8 quantization with no significant accuracy loss to report. 
\section{Applications}
\label{sec:apps}
\subsection{RF Device Identification and Authentication}

\begin{figure}[h]
\centering
\includegraphics[width=0.99\columnwidth]{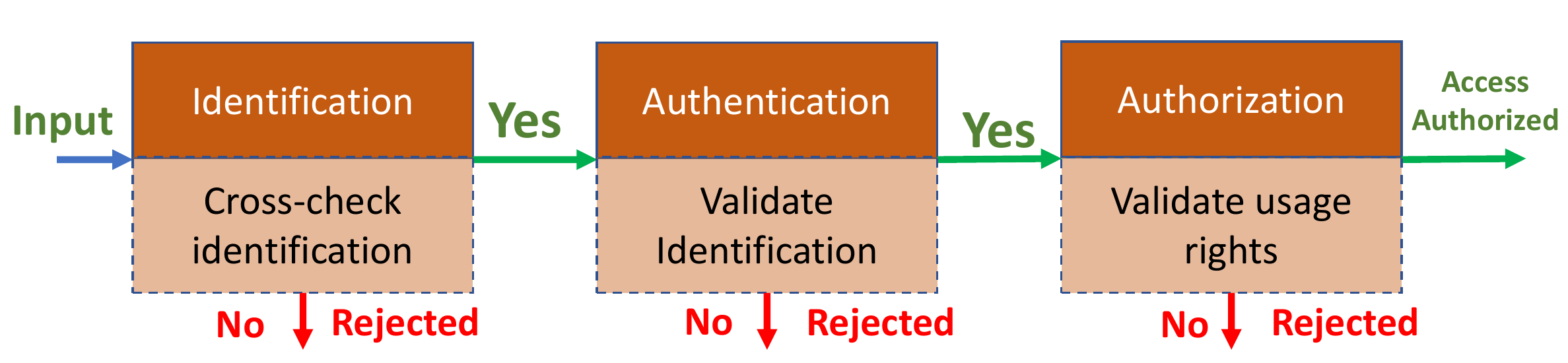} 
\caption{Flowgraph for Identification, Authentication and Authorization}
\label{fig:Authentication}
\end{figure}

Device identification and authentication are essential parts of managing wireless network. The proliferation of wireless devices in our environment is making this a daunting task due to the ever growing attack surfaces in the context of the burgeoning IoT economy. It is also often the case that identification and authentication are inaccurately used interchangeably causing further confusion \cite{manning2016device}. First, we provide definition for identification, authentication, and authorization along with Figure \ref{fig:Authentication} to encompass the overall process. Identification can be seen as a subtask of the overall authentication and authorization process. The definitions are provided below \cite{IBMdef},

\begin{enumerate}
    \item Identification is the ability to uniquely identify a user or device based on a unique ID such as MAC address, IMEI (International Mobile Equipment Identity) or MEID (Mobile Equipment Identifier) for phones.
    \item Authentication is the ability to prove that a user/device is genuinely who that user or device claims to be.
    \item Authorization is the process of evaluating whether a authenticated user/device has legitimate permission to access a resource or service.
\end{enumerate}

Traditionally, authentication involves handshake process between the device that intends to gain access and the network component that verifies the authentication message to grant access \cite{bassey2020device}. For example, using a secret key $S$, Alice may transmit a message to Bob using cryptography checksum which is a function of the message and the secret key. Bob can use the function and key to verify the authenticity of Alice while an intruder who tried to modify the message of the function will not be authenticated unless the secret key has been compromised. While this is not the sole method used in the industry for authentication, it is a typical representative example to demonstrate that the traditional authentication approach is active, i.e., it involves control message exchange and depends on secret keys. As one can imagine this leads to increased overhead due to the active nature of the authentication process. While secret keys are used for authentication process, it is still a point of vulnerability that could be compromised allowing illegitimate users gaining access to the network. Since the RF fingerprinting is hardware-specific and often unintentional characterization imparted at the analog component level, it is hard to mimic. Therefore, it could be argued that RF fingerprinting could be incorporated into more robust identification and authentication \cite{bassey2020device}. We explore the vulnerabilities of RF fingerprinting in section \ref{sec:open}.

\subsection{Localization, Tracking, and Navigation} 

As RF fingerprinting gains fidelity and robustness, it could be extended to or integrated with other applications such as assisting with outdoor or indoor localization, navigation, and even tracking of specific verified emitters. This could be highly beneficial for tactical applications, law enforcement, and first responders. For example, in search and rescue applications victims or rescue operators could be uniquely tracked based on the unique RF fingerprint emitted by their devices. There are several other examples of applications where such tracking can be highly beneficial. For example, there has been interest from the National Institute of Justice in using RF fingerprinting for contraband wireless devices tracking in correctional facilities \cite{Tracking}. Similarly, in most cases, it is useful to track the warfighters during the mission to monitor their progress, instantaneous location, and provide assistance when required. First responders often encounter tough situations but in most cases rely on wireless communication devices. Thus, if these communication signals can be used to identify and track the first responders, it can greatly enhance the efficiency and safety of these operations. 

It is important to point out that there is an added advantage that no specific packets need to be emitted to help with the tracking since it can be done implicitly by overhearing the communication signals. This could decrease the overall overhead required for command and control of such operations. This technology like many others is a double-edged sword, one could imagine security vulnerabilities where modern devices could be identified by illegitimate entities and then used to track leading to privacy and security concerns. This implies there is a whole new emerging area of research and analysis that may aim at mitigating such security vulnerabilities. 

\subsection{Intrusion detection} 

We discussed some of the security vulnerabilities that could arise from the misuse of RF fingerprinting but at the same time, it is a powerful tool to detect intrusions and/or imitation attacks. With the proliferation of wireless devices ranging from 5G mobile devices to low-cost IoT devices, it is becoming difficult to secure the ever-expanding threat surfaces. While there are millions of devices they all depend on a few wireless protocols or standards such as 5G, WiFi, BLE, LoRa, among others essentially implying that there will also be several devices that transmit the same kind of signals. Therefore, it is becoming imperative to have the ability to distinguish between legitimate and illegitimate users \textit{on-the-fly} even if they transmit the same signals. More importantly, intruders often mimic or perform replay attacks. Just like traditional fingerprints enable some of the security systems to detect intruders, RF fingerprinting can serve the same purpose for commercial and tactical applications. For example, before a squadron is deployed into a mission, each of them could have RF fingerprint information of their fellow warfighters. In this way, each device will be able to alert the presence of an intruder who is not part of the signature database. There are many commercial buildings where unauthorized wireless devices are prohibited, in such commercial secure environments, the security officers could deploy a similar methodology where every approved device is registered using their RF fingerprint. Once the system is activated, the intruder detection system will be able to alert the operator of unauthorized transmission even if they resort to replay or imitation attacks. 

\subsection{Application Domains}
\begin{figure*}[h]
\centering
\includegraphics[width=1.99\columnwidth]{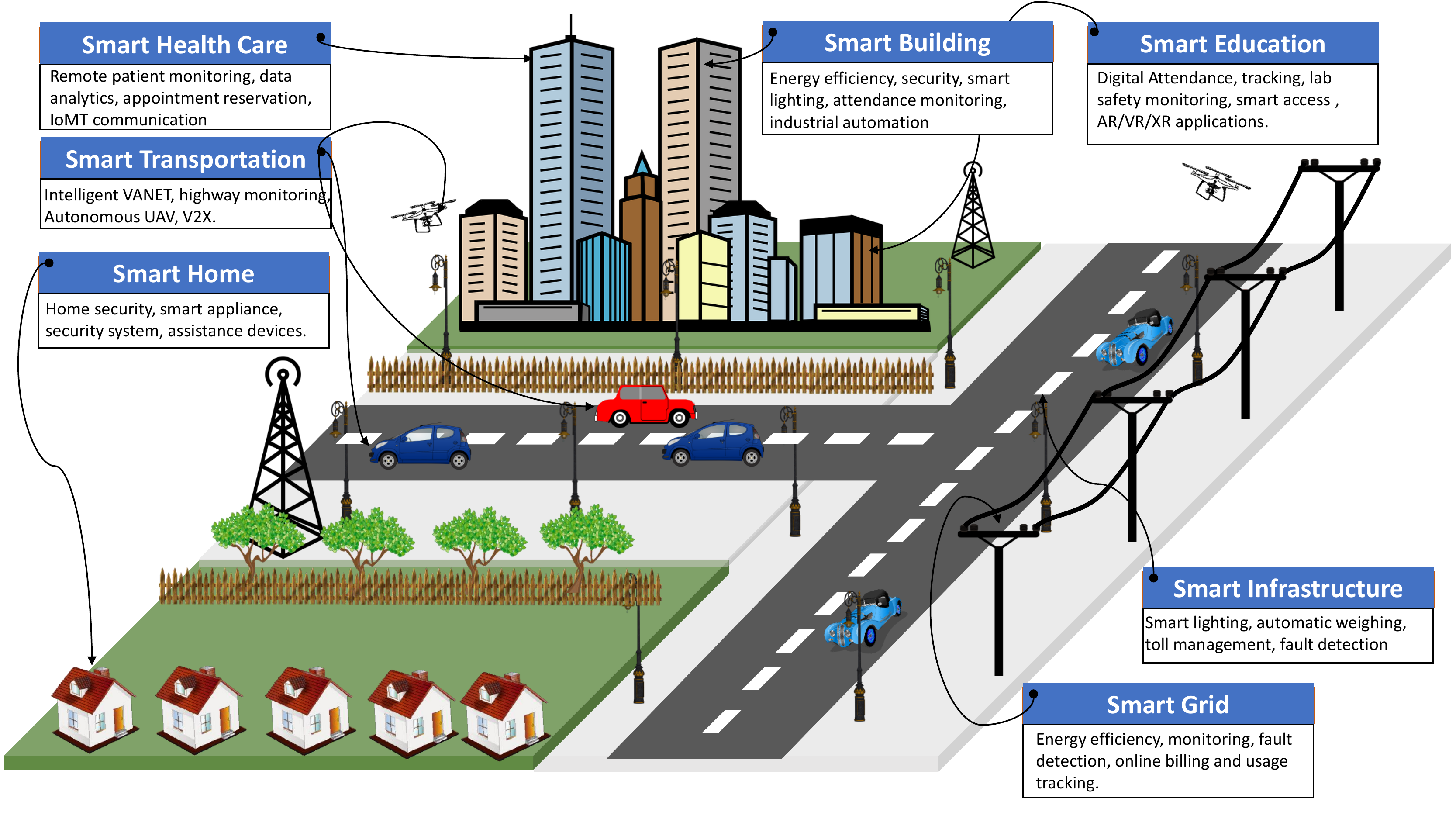} 
\caption{Application Domains for RF fingerprinting}
\label{fig:Application}
\end{figure*}

Beyond 5G network or 6G envisions revolutionary application domains \cite{6gwireless,Ajagannath6G2020,6g_road,6gaerialBS,Nayak20216GCT}. However, with such immersive applications security and privacy of users as well as assets become paramount \cite{wu2018survey, Ramezanpour22Comnet, ahmad2019security,RamezanpourZTA22, JagannathWiseML22_DT}. In this section, we shed some light on the envisioned applications for RF fingerprinting in the context of 6G as shown in Figure. \ref{fig:Application}.
\begin{enumerate}
    \item \emph{Intelligent Telehealth}: Intelligent and real-time healthcare will witness a paradigm shift with the 6G network. Real-time health monitoring, hospital-to-hospital services, Internet-of-Medical-Things (IoMT) also known as Healthcare IoT will collectively present dynamic and responsive health services \cite{Nayak20216GCT,SDNHealth}. Body area networks (BAN) with interconnected IoMT will advance and personalize telehealth monitoring and management. Remote health services with holographic teleconferencing with the ultra low latency 6G communication holds immense potential in democratizing healthcare services. Security and privacy for such an interconnected healthcare system that maintains the patient database and vital healthcare provider information is the backbone in realizing the tactile 6G healthcare. Wireless device fingerprinting solutions that resides on edge IoMT devices will be key to the real-time secure 6G IoT-based healthcare. We foresee that such solutions in conjunction with distributed ledger based multifactor authentication can secure the integrity and privacy of the users from spoofing, denial-of-service (DoS), identity theft attacks, among others.
    \item \emph{Autonomous UAV and V2X}: Aerial base stations and swarms of UAV can revolutionize and democratize wireless connectivity. Especially, setting up infrastructureless networks can provide emergency response, healthcare services, etc., by connecting remote and austere locations. Such concepts have been explored in the past \cite{Jagannath19ADH_HELPER} and will be a potential 6G use case \cite{6g_road,6gaerialBS,6gUAV}. Similarly, connected autonomous vehicles as in a V2X scenario would involve the vehicles communicating with nearby networks along its route. In these applications, handover from one network to another based on location and mobility is a necessity. Accordingly a robust, fast, and lightweight device authentication will be a key enabler to account for the diversity and mobility of devices accessing the network. RF fingerprinting which inherently involves no control overhead and merely uses hardware signatures embedded in the unintentional emissions will be an ideal candidate for such lightweight authentication schemes.
    \item \emph{Smart Grid 2.0} Smart grids are IoT-based electrical network for remote monitoring and control of power systems. With the advent of 6G, the smart grids will be able to support higher density of IoT devices for ultra low latency and high reliability communication enabling real-time anomaly detection and mitigation over distributed grid lines and stations. Confidentiality of the information managed in these power grids pertaining to user information, power metering, electrical usage patterns, billing details, among others are indispensable and primary targets of cybersecurity attacks. Moreover, smart grid 2.0 envisions intelligent pricing, automated grid management including energy trading among unknown parties in a point-to-point manner which further exposes the threat surfaces \cite{6g_road}. Device fingerprinting based authentication and grid access for secure energy trading will therefore gain popularity to realize smart grid 2.0.
    \item \emph{Extended Reality}: Extended reality (XR) is a blanket term to refer all real and virtual environments including virtual reality (VR), augmented reality (AR), mixed reality, and everything in between \cite{6gxr,6gstudy}. 6G will support advanced XR for various use cases such as military tactical training, video conferencing, online gaming, etc. In such applications along with meeting the latency and rate requirements, user privacy will be an equally necessary and challenging prerequisite. Consequently, user (device) authentication and access control will play a pivotal role here.
\end{enumerate}

\section{Traditional Approaches for RF fingerprinting}
\label{sec:traditional}
\subsection{Modulation domain based Approach}

\subsubsection{PARADIS}
The authors in \cite{Brik_ACM_2008} propose a radiometric signature-based device identification called PARADIS (Passive RAdiometic Device Identification System). This approach is based on the concept of radiometric identity - taking advantage of minor variations of transmitter hardware leading to peculiar features in the transmitted signal - to identify the origin. The authors demonstrate the accuracy of PARADIS to be greater than $99\%$ for classifying more than $130$ 802.11 Network interface cards (NICs). The system quantifies the transmitter's radiometric identity by comparing the signal with an ideal signal in the modulation domain on a frame-by-frame basis.

Modulation domain metrics such as frequency error, SYNC correlation, IQ offset, magnitude error, and phase error are used as the features for determining the radiometric identity of the device. The features resulting from hardware imperfections will be apparent over multiple frames. Therefore, calculating the statistical averages of these variations over multiple frames will magnify the artifacts caused by the hardware while at the same time reducing the effects of noise and channel. Following this, these five modulation domain metrics are classified using a classifier to identify the source. Two radiometric signature classifiers are implemented and evaluated, one using the SVM algorithm and the other using the k-nearest neighbors (k-NN) algorithm. The SVM classifier is built using LIBSVM \cite{Chang_libsvm}, the model takes a single radiometric signature as input and outputs the most likely identity of the source with the measure of confidence. A k-NN classifier is implemented using a group of rankers, where each NIC has one ranker that calculates the similarity between a given signal and the template of its signature computed during training.

The authors evaluated PARADIS on the ORBIT indoor wireless testbed facility \cite{orbit}. They collected data from 138 Atheros NICs configured as 802.11b access points on channel 1. Agilent 89641S vector signal analyzer was used as the PARADIS sensor to capture the frames from the transmitters. Overall, PARADIS using the SVM algorithm had an error rate of $0.0034\%$, and the system using the k-NN algorithm had an error rate of $3\%$ in classifying $138$ identical NICs.

\subsubsection{IQ Imbalance}
In \cite{zhuo2017iqImbalance}, the authors proposed a method to extract RF fingerprint features based on the IQ imbalance of the quadrature modulation signals. The IQ imbalance is caused due to the hardware imperfections in the IQ quadrature modulator. In the proposed method, the features are extracted by performing autocorrelation on the received signals. Real and imaginary parts of the autocorrelation form the RF fingerprint feature, and the SNR is estimated using the traditional least squares algorithm. To evaluate the method, the authors simulate five analog modulators (emitters) by varying the gain and orthogonal IQ imbalance and generate 400 signals from each emitter. The fingerprint feature vector is extracted using the proposed method and an SVM classifier is trained using half of the dataset. This method performs with an accuracy greater than $90\%$ for SNR $\geq15$dB and greater than $99\%$ for SNR $\geq20$dB.

\subsubsection{Modulation shape and spectral features}
The work in \cite{danev2009physical} proposes a method for identifying Radio Frequency Identification Devices (RFID) by extracting the RF fingerprint from modulation shape and spectral features of the signal emitted by the transponder when subjected to an RFID reader. The proposed method is able to identify $50$ identical RFID transponders from the same manufacturer with an error rate of $2.43\%$.

The authors use a purpose-built RFID reader to transmit to the target transponder for capturing the signals. It consists of two signal generators for envelope and modulation generation and a PCB antenna to transmit to the RFID transponder. The response from the RFID transponder is captured using an antenna and oscilloscope. Using this setup, the authors collected data from 50 JCOP NXP 4.1 smart cards and 8 electronic passports via the following four methods;
\begin{itemize}
    \item \textbf{Method 1}: Capturing the response of the transponder when subjected to ISO/IEC 14443 standard Type A and B protocols.
    \item \textbf{Method 2 (Varied $F_c$)}: Capturing the response of the transponder when subjected to out of specification (carrier frequency only) ISO/IEC 14443 standard Type A and B protocols.
    \item \textbf{Method 3 (Burst)}: Capturing the response of the transponder when subjected to bursts of RF energy (10 cycles of non-modulated $5$ MHz carrier at $10$V peak-to-peak).
    \item \textbf{Method 4 (Frequency sweep)}: Capturing the response of the transponder when subjected to linear sweep of a non-modulated carrier from $100$ Hz to $15$ MHz (at $10$ V peak-to-peak).
\end{itemize}
Modulation-shape features for data captured using methods 1 and 2 and spectral principal component analysis (PCA) for methods 3 and 4 are extracted. Modulation-shape features are extracted by performing Hilbert transformation on the captured signals. The starting point of the modulation in the transformed signal is located using a variance-based threshold detection algorithm \cite{Bonne2007}. Standardized Euclidean distance is obtained by matching the extracted fingerprint feature with the reference fingerprint \cite{manly2016multivariate}. Similarly, the Mahalanobis distance is evaluated by matching the reference fingerprint features to the test features. The spectral PCA features are extracted using a modified PCA for higher dimension data \cite{bishop_2006}.

To evaluate the classification capabilities of the proposed techniques, the authors consider signals captured from 8 e-passports and 50 JCOP NXP 4.1 cards. Both classification techniques, one using modulation features and the other using spectral features, perform with an error rate of $0\%$ when classifying signals into three classes (two countries, JCOP NXP card). The authors evaluate the identification capabilities of the method by using data collected through methods 3 and 4 consisting of data from 50 identical JCOP NXP 4.1 cards. The proposed method performs with an accuracy of $95\%$ in identifying 50 RFID cards when spectral features from data collected through methods 3 and 4 are used individually. Finally, when the spectral features from data collected through methods 3 and 4 are combined, the accuracy increases to $97.5\%$.


\subsubsection{Weighted Voting-Based Classification of Modulation Domain Signals}
In \cite{candore2009robust_radiometric}, the authors propose the use of a committee of weak classifiers to provide a strong classification by using weighted voting to combine results of multiple weak classifiers. Physical characteristics of the radio like frequency offset, modulation phase offset, in-phase/quadrature-phase offset, and magnitude are extracted from signals generated by six different radios in Wireless Open-Access Research Platform (WARP). Differential Quadrature phase-shift keying (DQPSK) modulation signals are generated and transmitted by the six radio cards. A total of 14 ML classifiers are built using the following signal characteristics:
\begin{itemize}
    \item Frequency difference (1 classifier): The distance between the actual transmission and ideal carrier frequency.
    \item Magnitude difference (4 classifiers): The distance between the magnitude of  transmitted  and ideal carrier symbols.
    \item Phase difference (4 classifiers): The angular distance between the transmitted and the ideal symbol in the IQ domain.
    \item Distance vector (4 classifiers): Vector distance between the transmitted and the ideal symbol.
    \item IQ origin offset (1 classifier): Distance between the origin of the ideal IQ plane and the origin of the transmitted symbol in the IQ domain.
\end{itemize}
The 14 classifiers are trained with the first 200 frames of 1844 random QPSK symbols from each board, then the outputs of the classifiers are combined using weighted voting to get the final radio identity. The weighted voting-based classifier has an average accuracy of $88\%$ in detecting six radio cards.

\subsubsection{Constellation Error Features}
The authors in \cite{huang2012constellation_error} propose an RF fingerprinting approach based on constellation error features. Transmitter imperfection that is reflected in error between the received constellation and the ideal constellation is used as the feature for RF fingerprinting. These features are extracted using subclass discriminant analysis (SDA). Burst QPSK modulated signals from seven TDMA satellite terminals are captured to construct the dataset for testing. The received signal is synchronized for time and frequency before building the modulation constellation, following which the constellation errors are computed. Feature vectors containing 41 features are extracted and classified for each of the signals using SDA feature extraction. The proposed method performs with an accuracy greater than $95\%$ for the bin size of the SDA feature extraction method greater than 12.

As we conclude this section, we summarize the reviewed literature in Table \ref{tab:modulation_label} for easy comparison for the reader.

\begin{table*}[h!]
    \centering
    \begin{tabularx}{0.9\textwidth}{>{\raggedright\arraybackslash}X >{\raggedright\arraybackslash}X >{\raggedright\arraybackslash}X >{\raggedright\arraybackslash}X >{\raggedright\arraybackslash}X >{\raggedright\arraybackslash}X }
         \hline
         Work & Radiometric Parameter & Classification technique & RF emitters & Performance\\
        \hline
        \hline
         \emph{Brik et al.}\cite{Brik_ACM_2008}& frequency error, SYNC correlation, IQ offset, magnitude error, and phase error & k-NN \& SVM & $138$ 802.11 NICs & SVM$\rightarrow99.9\%$ k-NN$\rightarrow97\%$ \\\\
         \emph{Zhuo et al.}\cite{zhuo2017iqImbalance}& IQ Imbalance & SVM & MATLAB simulated 5 analog modulators & $\geq90\%$ for SNR $\geq15$dB \\\\
         \emph{Danev et al.}\cite{danev2009physical}& Modulation-shape and spectral PCA features & Mahalanobis distance & 8 e-passports and 5 JCOP NXP 4.1 cards & $100\%$ classification accuracy and $97.5\%$ identification accuracy. \\
         \emph{Candore et al.}\cite{candore2009robust_radiometric} &frequency
offset, modulation phase offset, in-phase/quadrature-phase offset,
and magnitude & weighted voting-based classifier &Six WARP radio cards &88\% identification accuracy and 12.8\% false alarm rate\\
\emph{Huang et al.} \cite{huang2012constellation_error} &constellation-error &SDA
feature extraction &seven TDMA satellite terminals &95\% identification accuracy\\
        \hline
    \end{tabularx}
    \caption{Modulation-domain based RF fingerprinting works}
    \label{tab:modulation_label}
\end{table*}

\subsection{Statistical Approach}
\subsubsection{Non-Parametric Feature}
In \cite{patel2015non_parametric}, the generation and use of non-parametric features like mean, median, mode, and linear model coefficients (slope and intercept are estimated by linear regression) for identifying ZigBee devices is proposed. Complex IQ signals from four Texas Instruments ZigBee CC2420 devices are captured using Agilent E3238S receiver. The phase variable of the received signal is generated and the preamble region of the phase variables is divided into 32 equal sized Regions of Interest (ROIs). Following this, the non-parametric features are generated for each ROI. The signals are classified using a random forest classifier with 1000 trees. Each of the four non-parametric features is used individually to classify the device. The results show that the classification accuracy for each of the non-parametric features is above $97\%$ for SNR$\geq10$dB. At lower SNR, linear model coefficient features perform better than the other non-parametric features. The author also compares the performance of using the non-parametric features over parametric features by computing the parametric features (variance, skewness, and kurtosis) for each ROI. The same random forest classifier is used to classify the features individually. The non-parametric features show improvements by upto $9\%$ at SNR$=8$dB over the parametric features.

\subsubsection{RF-DNA based features}
In \cite{lukacs2015rf_dna}, the authors propose a RF distinct native attribute (RF-DNA) based RF fingerprinting for identifying ultra-wideband (UWB) noise radar emitting devices. RF-DNA fingerprint features, including variance, skewness, and kurtosis, are extracted for the time-domain response of the signals. Additionally, the authors also extract normalized power spectral density (PSD) and discrete Gabor transform from the spectral-domain response of the signals. The signals are classified using multiple discriminant analysis with maximum likelihood (MDA/ML) classifier and generalized relevance learning vector quantization-improved (GRLVQI) classifiers. MDA/ML classifier is a combination of multiple discriminant analysis (MDA) that aims to reduce the dimensionality of a multi-dimensional dataset and maximum likelihood (ML) classifier. The GRLVQI classifier is a supervised machine learning method that enlarges generalized learning vector quantization (GLVQ) by adding weighting factors to the input dimensions \cite{hammer2002GRLVQ}. These factors allow for appropriate scaling of the input dimensions according to their relevance and are adapted automatically during training according to the specific classification task.
\\
The classification performance is evaluated on signals captured by transmitting UWB noise radar waveforms using a log-periodic antenna placed one meter from the receiver in an anechoic chamber. By varying the termination load of the transmitting antenna, three classes of waveforms are captured. Additionally, an attenuator is used to increase the number of classes. For the three-class case using only time-domain fingerprint features, the proposed method has a classification accuracy of $99.7\%$ and $98.25\%$ for MDA/ML and GRLVQI classifiers, respectively. In the case of the seven-class dataset using only time-domain fingerprint features, the proposed method has an average classification accuracy of $81\%$ and $75\%$ for MDA/ML and GRLVQI classifiers, respectively. Similarly, for the three-class dataset using only spectrum-domain fingerprint features, the proposed method has a classification accuracy of $91.97\%$ and $94.47\%$ for MDA/ML and GRLVQI classifiers, respectively. Lastly, using a combination of time and frequency domain features, the classification accuracy for the three-class data is $97.65\%$ and $93,79\%$ for MDA/ML and GRLVQI classifiers, respectively.

\subsection{Transient-based Approach}
Transient-based approach involves identifying distinctive features present in the radio turn-on transients, which appear at the start of the transmission. The transient is the section of the signal where the amplitude rises from channel noise to signal amplitude. Identification of devices using this process consists of three steps: detection of transients, extraction of features, and classification. A brief overview of the two key approaches of transient detection: Threshold\cite{shaw1997threshold} and Bayesian step-change detector\cite{ureten1999detection} is discussed in \cite{hall2003detection}. Both of which exploit the amplitude characteristics of the signal for transient detection. They also propose a new approach for transient detection using the phase characteristic of the signal to improve performance when the transient gradient is gradual.

\subsubsection{Fast Fourier Transform (FFT)-based Fisher features}
The authors of \cite{danev_transient} propose the use of FFT-based Fisher features to identify wireless nodes. In this approach, the RF fingerprint (feature template) of the signals is computed by first detecting the start point and extracting the transient of the signals using a variance-based threshold detection algorithm \cite{Bonne2007}. The relative difference between the adjacent FFT spectra is determined by applying a 1-D Fourier Transform on the transients. Following which the Fisher feature vector that forms the feature template is extracted using a Linear Discriminant Analysis (LDA) matrix. The LDA matrix is derived by a standard procedure based on scatter matrices \cite{martinez_lda}. Lastly, the fingerprint is matched by calculating the Mahalanobis distance between the reference and test signals feature template.

Over 600 IEEE 802.15.4 signal samples from 50 consumer-off-the-shelf (COTS) Tmote sky sensor nodes with the same manufacturer signature are collected to evaluate the proposed technique. The system identifies the 50 sensors with an accuracy higher than $99.5\%$. This work also investigates the effects of parameters such as distance, antenna polarization, and voltage on the performance of the system. 

The results of these investigations suggest the system is robust against distance, multipath propagation, and voltage changes. But a change in the polarization of the signal alters the shape of the transient perturbing the frequency information present in the transient consequently leading to a drop in recognition accuracy. They also investigate the practicality of the system on attacks such as hill-climbing and DoS. Hill-climbing attack is a common impersonation attack where the attacker repeatedly submits data to the algorithm with slight modification. Modifications that improve or preserve the matching score are preserved. Over time, the attacker can achieve a score higher than the designed threshold resulting in a successful impersonation. The system can be vulnerable to an impersonation attack when the number of signals used to build the fingerprint feature template is low. This work investigates the vulnerability of the system to jamming-based DoS attacks. Due to the superposition of the original and the jamming signals, the system is unable to recognize the device. The authors suggest that this type of attack can be used as a security measure against an attacker.

\subsubsection{Hilbert-Huang transform-based time-frequency-energy distribution features}
The authors propose a specific emitter identification (SEI) method based on the transient signal's time-frequency-energy distribution (TFED) obtained by Hilbert-Huang transform (HHT) \cite{yuan2014HilbertHuang}. HHT is a self-adaptive signal analysis method it involves Empirical Mode Decomposition (EMD) and Hilbert transform \cite{huang1998empirical}. The EMD method decomposes a given signal into a set of a finite number of intrinsic mode functions (IMFs). Applying Hilbert transform on the IMFs yields the TFED. The start of the signal is detected using a phase-based method \cite{hall2003detection} and the endpoint is detected by forming the energy trajectory of the signal from TFED. Following thirteen features are extracted from the TFED:
\begin{itemize}
        \item Three features from overall features: Sum of energy, duration of transient signal, and duration of the maximum energy point.
        \item Four energy distribution features along the frequency-axis: entropy, kurtosis, skewness, and center.
        \item Four energy distribution features along the time-axis: entropy, kurtosis, skewness, and center.
        \item Two energy distribution features of the overall time-frequency plane: entropy and center.
\end{itemize}
The authors use PCA to reduce the dimension of the feature vector and use an SVM to classify the devices. The authors collect transient signals from eight GSM mobile phones (four Nokia 5230, two Motorola Me525, and two Xiaomi-1) using a Leroy 8500A digital oscilloscope connected to a digital receiver with a Yagi antenna. The SVM classifier is trained with 50 transient signals from each device, and the system is tested with 100 transient signals from each device. The proposed method attains an accuracy of $100\%$ in classifying the eight mobile devices.
  
\subsubsection{Energy envelope features}
In \cite{rehman2012energy_envelop}, features extracted from the energy envelope of the transient signal are used as RF fingerprints to identify Bluetooth devices. The transients are extracted from the normalized signals using variance-based threshold method \cite{Bonne2007}. The energy envelope is then extracted using the spectrogram which is defined as the squared magnitude of Short-Time Fourier Transform (STFT). The spectrogram computes a three-dimensional TFED that is then sliced with respect to the instantaneous frequency at the maximum energy value to obtain the smoothed energy envelope curve. Finally, the RF fingerprint is formed by extracting the following features from the energy envelope curve:
\begin{itemize}
        \item Area under the normalized curve.
        \item Duration of transient.
        \item Maximum slope of the curve.
        \item Kurtosis of the curve.
        \item Skewness of the curve.
        \item Variance of the transient envelope.
\end{itemize}
  
Bluetooth device discovery mode signals from seven built-in Bluetooth transceivers of cell phones are captured using an oscilloscope and an Agilent spectrum analyzer. A total of 300 signals from each of the seven Bluetooth transceivers are captured and the RF fingerprint features are extracted. A k-NN classifier with $3$ nearest neighbors is used to classify the feature vector. Fifty signals from each of the seven devices are used to train the classifier, and the remaining 250 signals are used for testing. The proposed method classifies the devices with an accuracy of $99.9\%$. Further, the authors investigated the effect of sampling rate on classification accuracy. Because the energy envelope of the transient does not change with the sampling rate, the accuracy of identifying devices remains $99.9\%$ for a sampling rate of 4 GSps, 1 GSps, 512 MSps, 256 MSps, 128 MSps, and 32 MSps.
  
The above discussed transient-based RF fingerprinting literature are also summarized in Table \ref{tab:transient_label}.
\begin{table*}[h!]
    \centering
    \begin{tabularx}{0.9\textwidth}{>{\raggedright\arraybackslash}X >{\raggedright\arraybackslash}X >{\raggedright\arraybackslash}X >{\raggedright\arraybackslash}X >{\raggedright\arraybackslash}X >{\raggedright\arraybackslash}X }
         \hline
         Cited literature & Transient detection method & Classification technique & RF emitters & performance \\
        \hline
        \hline
         \emph{Danev et al.}\cite{danev_transient}& Variance-based threshold & Mahalanobis distance & 50 COTS Tmote Sky sensor (IEEE 802.15.4) & $\geq99.5\%$ \\\\
         \emph{Yuan et al.}\cite{yuan2014HilbertHuang} & Phase-based method & SVM & 8 GSM Mobile phones & $100\%$ \\\\
         \emph{Rehman et al.}\cite{rehman2012energy_envelop} & Variance-based threshold & k-NN & 7 built-in Bluetooth transceivers & $99.9\%$ \\
        \hline
    \end{tabularx}
    \caption{Transient based RF fingerprinting works}
    \label{tab:transient_label}
\end{table*}

\subsection{Wavelet-based approach}
\subsubsection{Dual-tree complex wavelet transform (DT-CWT)}
A wavelet domain (WD) approach based on DT-CWT features extracted from the non-transient preamble response of 802.11a signals is proposed by the authors of \cite{klein_wavelet}. The effectiveness of WD fingerprinting is demonstrated using Fisher-based MDA/ML classification. Also, this work considers the effect of varying channel SNR, burst detection error, and dissimilar SNRs for MDA/ML training and classification. WD fingerprinting with DT-CWT features achieves classification accuracy of $80\%$ for signals with SNR up to 8 dB and performs superior to time-domain RF fingerprinting.

\begin{figure*}[h]
\centering
\includegraphics[width=5.8 in]{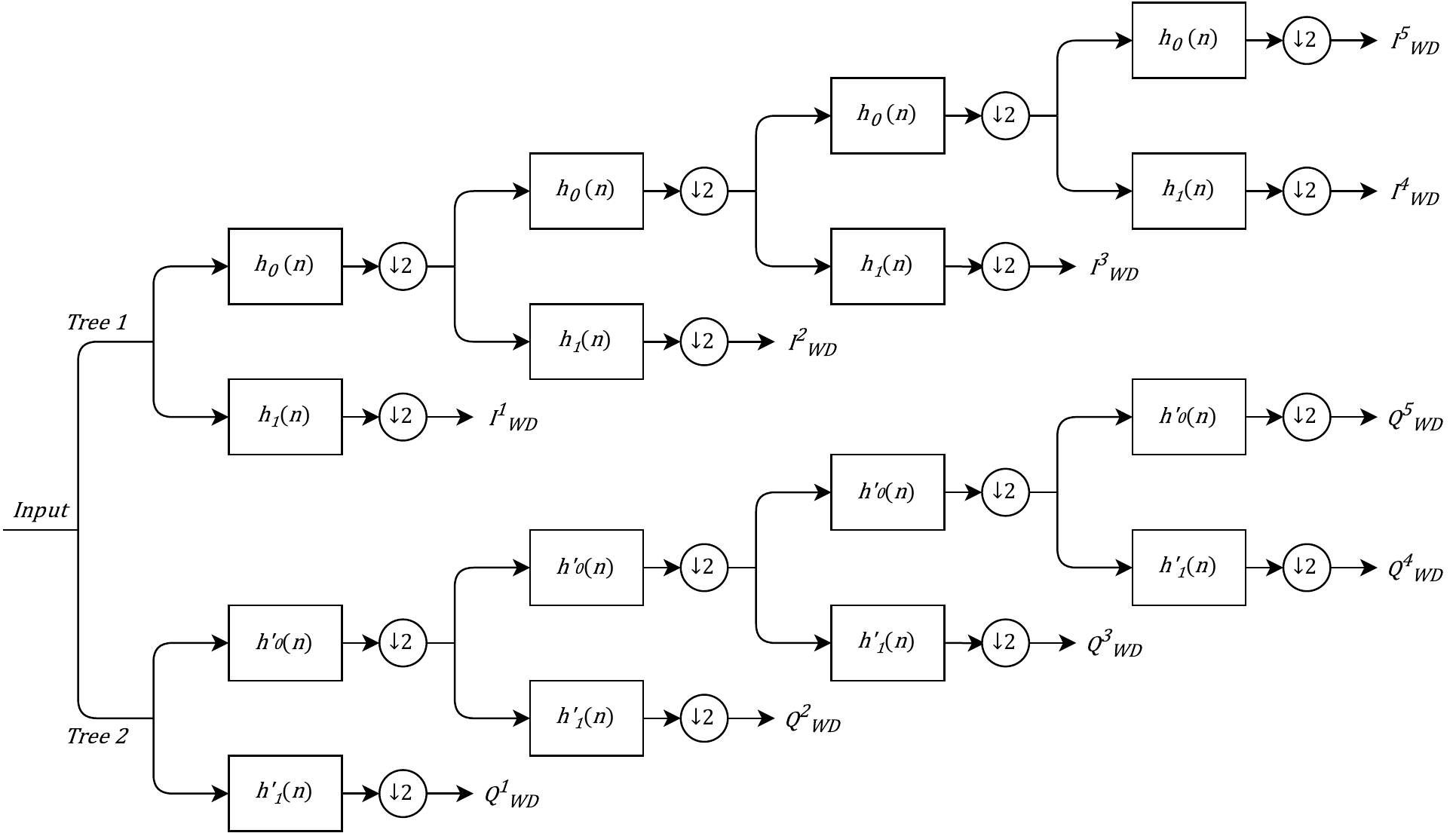}
\caption{Five level dual-tree complex wavelet transform}
\label{fig:dt-cwt}  
\end{figure*}

DT-CWT is an extension of discrete wavelet transform (DWT) which decomposes a time-domain signal into wavelets that are localized in frequency and time domain. DT-CWT addresses the necessity of shift-invariance that is not present in DWT. DT-CWT is implemented using two real-valued filter banks represented as \emph{tree1} and \emph{tree2} in Figure \ref{fig:dt-cwt} \cite{selesnick_dt-cwt}. The wavelet and scaling functions for \emph{tree1} filter banks - $\psi(t)$ and $\phi(t)$ - are given by,
\begin{equation}
    \psi(t) = \sqrt{2}\sum_{n}h_{1}(n)\phi(2t-n) \label{eq:dt_cwt psi},
\end{equation}
\begin{equation}
    \phi(t) = \sqrt{2}\sum_{n}h_{0}(n)\phi(2t-n) \label{eq:dt_cwt phi}
\end{equation}
and the corresponding functions for \emph{tree2} filter banks are Hilbert transforms of Equations (\ref{eq:dt_cwt psi}) and (\ref{eq:dt_cwt phi}) given by,
\begin{equation}
    \psi'(t) = \sqrt{2}\sum_{n}h'_{1}(n)\phi'(2t-n) \label{eq:dt_cwt psi'},
\end{equation}
\begin{equation}
    \phi'(t) = \sqrt{2}\sum_{n}h'_{0}(n)\phi'(2t-n). \label{eq:dt_cwt phi'}
\end{equation}
The filter coefficients $h_{1}(n), h_{0}(n), h'_{1}(n)$, and $h'_{0}(n)$ are implemented directly as analysis filters. For a real-valued input, the DT-CWT filter bank outputs a real-valued WD component $I_{WD}^{l}$ and an imaginary component $Q_{WD}^{l}$. From this, a complex WD signal can be expressed as,
\begin{equation}
    s_{WD}^{l}(n) = I_{WD}^{l}(n) + jQ_{WD}^{l}(n). \label{eq:dt-cwt_complex}
\end{equation}
To mitigate the excessive need for computation time and data processing when using fundamental signal characteristics such as amplitude $\alpha(n)$, phase $\phi(n)$, and frequency$f(n)$, as the classification features. The authors propose to use the statistical properties of the fundamental signals for the classification of devices. These statistics include variance, skewness, and kurtosis.

The authors collect the IQ samples from laptops with 802.11a Cisco personal computer memory card international association cards using an Agilent-based RF signal intercept and collection system in an anechoic chamber. To simulate the various SNR conditions, an "analysis signal" is generated by adding a random complex additive white Gaussian Noise (AWGN) signal to the collected complex signal. Before adding, the noise signal is filtered and power-scaled to achieve desired SNR for the analysis signal. Next, the starting location (sample number) of the RF burst is visually determined and is used to locate the preamble region. The analysis signals are divided into three subregions for fingerprint generation. A five-level DT-CWT is performed for each of these subregions, and the complex WD signal for each of the levels is computed for all the subregions using (\ref{eq:dt-cwt_complex}), followed by the calculation of signal characteristics and statistical classification features 
resulting in a total of 135 features per analysis signal, which is then used for Fisher-based MDA/ML classification with Monte Carlo simulation and $K$-fold validation.

To compare the proposed WD fingerprints with time-domain (TD) fingerprints, the authors generated WD and TD fingerprints for each analysis signal. The TD fingerprints are generated similarly to WD fingerprints but without performing DT-CWT, which consists of three signal characteristic features and three statistical features for each of the three subregions. For each analysis signal, TD fingerprints are composed of $27$ features, and WD fingerprints are composed of $135$ features. Both techniques performed identically for SNR $\geq 25$ dB and the WD technique was superior for $-2 < $SNR$ < 24$ dB. WD fingerprints achieves an accuracy of $80\%$ at SNR $\approx 11$ dB. This performance increase when using WD fingerprinting is a gain of approximately 7 dB with respect to equivalent TD fingerprinting. To evaluate whether the classifier takes advantage of the larger number of features in WD fingerprints, the authors decided to choose a subset of 27 selected WD features from the 135 features. Classification with 27-feature WD fingerprinting shows that the WD technique outperforms the 27-feature TD fingerprinting only for $0 <$ SNR $< 20$ dB with a performance gain of approximately 2 dB relative to 27-feature TD fingerprinting. This increase in performance, given equal dimensionality, suggests that the classifier exploits the additional information present in DT-CWT features. 

\subsubsection{Dynamic wavelet}
A dynamic wavelet fingerprint method to identify unique RFID tags using supervised pattern classification techniques is presented in \cite{Bertoncini2012rfid}. In this study, 146 individual RFID tags of three types: Avery-Dennison AD 612, Avery-Dennison Runway Gen 2, and Alien Omni-Squiggle, are used. RF signals from each of the tags are captured by writing the same code onto the tag using Thing Magic Mercury 5e RFID Reader and reading the response with an omnidirectional antenna through a vector signal analyzer. From the captured complex-valued signals, amplitude, phase, and instantaneous frequency are computed and used for extracting RF fingerprint feature vector. In this work, the authors propose using a feature vector that is a combination of features extracted from dynamic wavelet fingerprint (DWFP), wavelet packet decomposition (WPD), and higher-order statistics.

The authors use the DWFP technique \cite{hou2002DWFP} that applies wavelet transform on the original TD signal and generates a "fingerprint-like" binary image. Image processing routines are performed on the binary image to extract signal's RF fingerprint. Feature selection is performed on the features extracted by the image processing steps using Euclidean distance metric to indicate the most highly separable interclass distance. Next, fingerprint features are extracted by performing WPD \cite{Rachel1995WPD}, which is done by applying a wavelet packet transform (WPT) on the RF signal to generate a tree of coefficients. Wavelet packet energy is calculated for the terminal nodes of the WPT tree and the highest energy is selected as the feature. Finally, higher-order statistics are performed on the unfiltered waveform to extract the following features: mean of EPC, maximum cross-correlation with another EPC from the same tag, variance, Shannon entropy, second central moment, skewness, and kurtosis. A combination of the features extracted with the three methods is used as the feature vector for the classifier. The proposed feature vector is tested with four types of classifiers for identifying RFID tags: Linear and Quadratic discriminant classifiers (LDC and QDC), k-NN, and SVM. All of the four classifiers perform with an accuracy of $99\%$ in identifying the RFID tags. 

\subsubsection{Wavelet domain-based Bayes approach}
In \cite{Ezuma2019MicroUAV}, the authors propose a WD-based Bayes approach to detect the presence of micro-UAVs and signal energy transient to identify the type of micro-UAV. The proposed detection method first converts the RF signals from the UAV controllers into WD using three-stage wavelet decomposition followed by differentiating noise (non-UAV signals) and micro-UAV signals using a naive Bayes approach based on the Markov model. The transformation of RF signals to the WD removes the bias and reduces the size of the data. If micro-UAV is detected using the proposed method, the classification of the signal is carried out. For the proposed classification method, the TD RF signal is first transformed into the energy-time-frequency domain and is represented as a spectrogram. The spectrogram is the squared magnitude of the discrete STFT of the signal. Energy transient is estimated by detecting the abrupt change in the energy trajectory from the spectrogram. The energy transient is then used to extract the statistical RF fingerprints (feature set) such as skewness, variance, energy spectral entropy, and kurtosis. The dimensionality of the feature set is reduced using Neighborhood Component Analysis (NCA). NCA is a supervised learning method for feature selection, transforming the primary data into a lower-dimensional space \cite{goldberger2004NCA}. The reduced feature sets are used to train four machine learning algorithms: k-NN, discriminant analysis (DA), SVM, and neural networks (NN). 

RF signals from 14 micro-UAV controllers operating at 2.4 GHz are captured indoors using Keysight MSOS604A oscilloscope. An omnidirectional antenna is used to capture the RF signal at a close distance and a grid antenna is used for far-field signal capture. A total of 100 RF signals is captured from each micro-UAV controller to form the dataset, which is split randomly with a ratio of 4:1 for training and testing. The authors show that the proposed detection method has an accuracy of $84\%$ in detecting the presence of micro-UAV for a given SNR of 10 dB and $100\%$ accuracy for SNR$\geq12$ dB. Once the UAV is detected, the RF signal is classified to identify the UAV. The classification accuracy of k-NN, SVM, DA, and NN classification methods are $96.3\%$, $96.84\%$, $88.15\%$, and $58.49\%$, respectively. Accuracy of classification increases with an increase in SNR. The authors clearly state that the hyperparameters of the NN algorithm were not optimized in this work, leading to the poor performance of the NN algorithm. If the hyperparameters of the NN algorithm are optimized and tuned correctly, the NN algorithm could have a high classification accuracy.

These discussed wavelet-based approaches are tabulated in Table \ref{tab:wavelet_label}.
\begin{table*}[h!]
    \centering
    \begin{tabularx}{0.9\textwidth}{>{\raggedright\arraybackslash}X >{\raggedright\arraybackslash}X >{\raggedright\arraybackslash}X >{\raggedright\arraybackslash}X >{\raggedright\arraybackslash}X }
         \hline
         Work & Wavelet method & Classification technique & RF emitters & Performance\\
        \hline
        \hline
         \emph{Kelin et al}\cite{klein_wavelet}& dual-tree complex wavelet transform (DT-CWT) & Fisher-based MDA/ML & 802.11a Cisco PCMCIA cards & $80\%$ at 11db SNR and $\geq 98\%$ at SNR $\geq25$dB \\\\
         \emph{Bertoncini et al.}\cite{Bertoncini2012rfid}& dynamic wavelet fingerprint (DWFP)\cite{hou2002DWFP}, wavelet packet decomposition (WPD)\cite{Rachel1995WPD} & LDC, QDC, k-NN and SVM & 50 Avery-Dennison AD 612, 50 Avery-Dennison Runway Gen 2, and 50 Alien Omni-Squiggle & $99\%$ \\\\
         \emph{Ezuma et al.}\cite{Ezuma2019MicroUAV}& three-stage wavelet decomposition & k-NN, discriminant analysis (DA), SVM,and neural networks (NN) & 14 micro-UAV controllers & k-NN$\rightarrow96.3\%$, SVM$\rightarrow96.84\%$, DA$\rightarrow88.15\%$, and NN$\rightarrow58.49\%$  \\
        \hline
    \end{tabularx}
    \caption{Wavelet-based RF fingerprinting methods}
    \label{tab:wavelet_label}
\end{table*}

\subsection{Other Approaches}
\subsubsection{Steady State Frequency Domain Approach}
The authors of \cite{kennedy_FreqDomain} present a technique for radio transmitter identification based on frequency domain characteristics. This approach employs frequency domain analysis with a traditional discriminatory classifier - k-NN - for RF fingerprinting and device identification. This work demonstrates an accuracy of $97\%$ at $30$ dB SNR and $66\%$ accuracy at $0$ dB SNR in identifying eight identical USRP transmitters.  

For demonstration, the authors consider the Random Access Channel (RACH) preamble in UMTS. The IQ samples of the preamble are captured and down-converted from transmit band to baseband. The baseband signal is bandpass sampled by the analog-to-digital converter (ADC) at the Nyquist rate and downsampled using a sum of absolute values window function followed by carrier frequency offset correction and amplitude normalization. Spectral analysis of the entire preamble signal is performed using the FFT and is fed as the input for the k-NN classifier. The dataset is divided into training and testing sets. In the training step, the k-NN algorithm maps the training preamble signals set into a multidimensional feature space, divided into regions based on the class. During testing, the preamble is determined to belong to the class with the most frequent label among the k-nearest preambles from training.

To evaluate the method UMTS RACH preamble are generated using MATLAB and transmitted using USRPs with identical specifications. An Anritsu Signature MS2781A spectrum analyzer is used to capture the IQ samples from eight USRPs individually and $300$ preamble samples are captured from each of the eight USRPs. For training the k-NN algorithm, $150$ preamble samples from each USRP is used and the remaining is used for testing the system. The system achieves a classification accuracy of $97\%$ for preamble signals above $25$ dB SNR and accuracy of $66\%$ for $0$ dB SNR. The authors also test the effect of binning on classification accuracy by varying the number of bins used to determine the spectral energy features. At lower SNR, binning reduces the overall classification performance and the accuracy reaches its maximum at around $200$ bins for SNR $15$ dB - $30$ dB.

\subsubsection{Permutation Entropy}
In \cite{deng2017radio}, the authors propose a multidimensional permutation entropy-based RF fingerprinting method. Permutation entropy is the measure of complexity for a given time series. Accordingly, it can extract and amplify the minuscule changes in the given time signal. The proposed method involves first capturing the radio signals and extracting the envelopes of the signal, then calculating the multidimensional permutation entropy of the signal envelope to form the RF fingerprint feature vector. 
An SVM classifier with an radial basis function (RBF) kernel is used to classify the feature vector. To evaluate the method, the authors collect 100 sets of data from three AKDS700 radios using a digital receiver and an oscilloscope. Multidimensional permutation entropy is computed for all the signals captured using a multidimensional vector. 
The SVM trained for these features performs with an average accuracy of $90\%$ for SNR$\geq10$ dB in recognizing the three radios.

\begin{table*}[h!]
    \centering
    \begin{tabularx}{0.9\textwidth}{>{\raggedright\arraybackslash}X >{\raggedright\arraybackslash}X >{\raggedright\arraybackslash}X >{\raggedright\arraybackslash}X >{\raggedright\arraybackslash}X >{\raggedright\arraybackslash}X }
         \hline
         Work & Radiometric Parameter & Classification technique & RF emitters & Performance\\
        \hline
        \hline
         
\emph{Kennedy et al.} \cite{kennedy_FreqDomain} &FFT &k-NN &8 USRPs &97\% identification accuracy at SNR>25 dB and 66\% at 0 dB SNR\\
\emph{Deng et al.} \cite{deng2017radio} &multidimensional
permutation entropy &SVM &3
AKDS700 radios &90\% identification accuracy at SNR$\geq$10 dB\\
\emph{Yuan at al.} \cite{yuan2019mfmcf} &RSS, SSD, and HLF features &MFMCF &7 APs & probability of zero positioning error is 96.5\%.\\
\emph{Baldini et al.} \cite{baldini2017PNandDN} &Permutation entropy and Dispersion entropy &k-NN, SVM, and decision tree &9 nRF24LU1+ &k-NN up to 82.3\%, SVM up to 82.1\%, and Decision tree up to 81.4\%.\\
        \hline
    \end{tabularx}
    \caption{Other traditional RF fingerprinting works}
    \label{tab:other_label}
\end{table*}

\subsubsection{Received Signal Strength}

A Multi-Fingerprint and Multi-Classifier Fusion (MFMCF) localization method for RF fingerprinting is proposed in \cite{yuan2019mfmcf}. The proposed technique aims to increase the localization accuracy of WiFi Access Points (APs) by constructing composite fingerprints and combining multiple classifiers. The authors construct a composite fingerprint set (CFS) consisting of received signal strength (RSS), signal strength difference (SSD), and hyperbolic location fingerprint (HLF) features. In this method,  a decision structure with three classifiers k-NN, SVM, and random forest is used to obtain a more accurate location estimate.

The authors collect RSS data of seven APs at 35 points in an indoor location, each at least 1.2 meters apart. A total of 100 RSS data is recorded for each of the APs at each location. Grubbs method \cite{grubbs1969procedures} based on the mean and standard deviation is used to detect outliers in RSS data. The outliers are replaced with a Gaussian random number generated using the mean and variance of the non-abnormal data. SSD and HLF fingerprints are constructed based on the collected RSS. SSD is the difference in RSS values observed by two APs, and HLF is the ratio of RSS between pairs of APs. The three fingerprints, RSS, SSD, and HLF are combined to form the CFS. Linear discriminant analysis is used to reduce the dimensions of CFS. Using the reduced CFS, the three classifiers (K-NN, SVM, and random forest) are trained. In the testing stage, the entropy of each of the classifiers is calculated and the classifier with the least entropy is used to estimate the location.

To evaluate the proposed MFMCF technique, the authors use LDA to select 12 features from 49 features in CFS, which covers more than $95\%$ of the information. The probability of zero positioning error of MFMCF is $96.5\%$, which is an increase of $4.2\%$, $6.4\%$, and $7.7\%$ compared with RSS, SDD, and HLS, respectively, when used as independent fingerprint features for classification. To compare MFMCF with independent classifiers, CFS was used to train and test individual classifiers. The probability of zero positioning errors of MFMCF, RF, k-NN, and SVM were $96.5\%$, $90.2\%$, $92.9\%$, and $94.8\%$, respectively. The authors also show that the proposed MFMCF technique has the lowest average localization error of $0.14$m. 
\subsubsection{Permutation entropy and Dispersion entropy}
The authors in \cite{baldini2017PNandDN} propose an RF fingerprinting method for identifying IoT devices using entropy-based statistical features called Permutation Entropy (PE) and Dispersion Entropy (DE).  In this work, nine nRF24LU1+ IoT devices are used for evaluating the proposed method. The RF signals from these devices are captured using an N210 USRP with XCVR2450 frontend. All nine devices are configured to transmit fixed payloads based on MySensors specifications. MySensors \cite{mysensors} is a free and open-source software framework for DIY (do-it-yourself) wireless IoT devices that allows devices to communicate using radio transmitters. The real-valued IQ samples are captured using the USRP followed by synchronization and normalization to obtain the burst of traffic associated with the payload. 

The following statistical features are then computed for each received payload: variance, skewness, kurtosis, Shannon entropy, log entropy, PE (order=4, and delay=1), PE (order=5, and delay=1), DE  (embedding dimension=3, classes=5, and delay=1), DE (embedding dimension=4, classes=5, and delay=1), and DE (embedding dimension=5, classes=5, and delay=1). The authors train three classifiers: k-NN, SVM, and decision tree with a subset of the ten features listed above. The authors show that the classifier trained using PE and DE features along with statistical features has an accuracy of $24\%$ to $30\%$ higher than the classifier trained with just statistical features (Shannon entropy and log entropy). Using just the PE feature along with statistical features leads to a good improvement in accuracy in contrast to using Shannon entropy and log entropy. Finally, the authors show that all three classifiers performed with similar classification accuracy when trained with PE and DE features along with statistical features.

The works discussed in this section are also condensed in a tabular form in Table \ref{tab:other_label}.
\section{Deep Learning for RF fingerprinting}
\label{sec:dlrff}
Deep learning based techniques have been slowly invading this field of research and becoming the state of the art. This is primarily due recent revival of machine learning fueled from rapid growth of computational capabilities and the availability of digital data. Keeping that in mind and for the benefit of reader who might be relatively new to deep learning, we provide a brief tutorial regarding the core techniques used for RF fingerprinting. For a more comprehensive review we recommend the readers to \cite{Jagannath20UAVBook}. 

\subsection{Overview on Supervised Deep Learning}

\subsubsection{Feedforward Neural Networks}
\label{sec:fnn}
\begin{figure}[h]
\centering
\includegraphics[width=3.5 in]{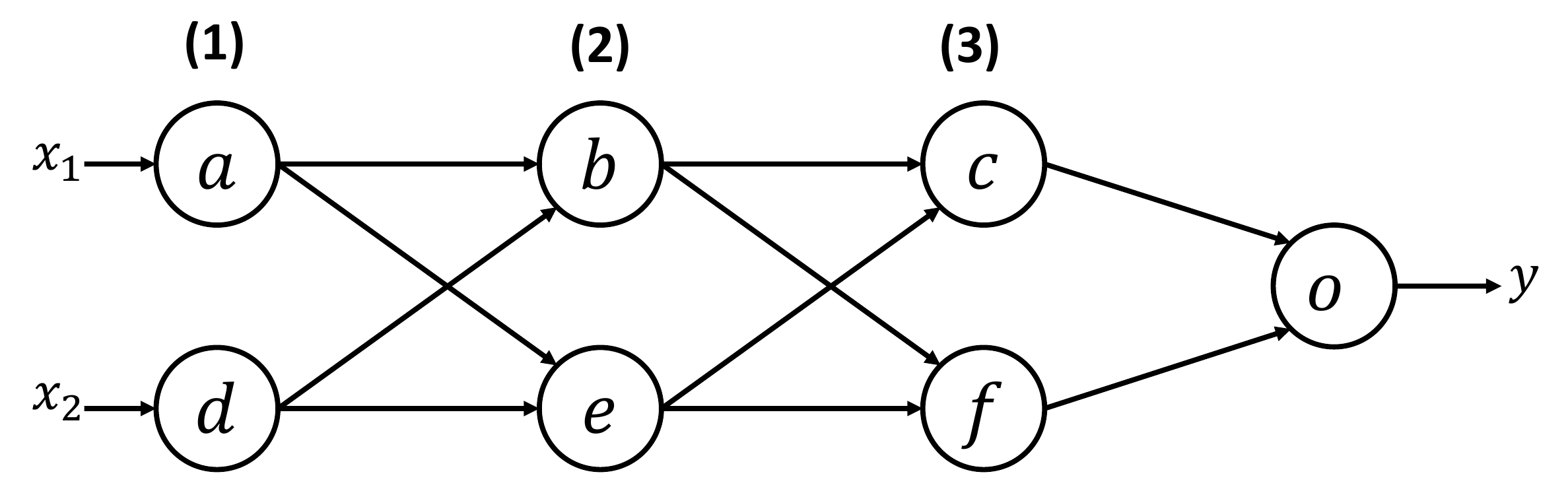}
\caption{Three-layered FNN}
\label{fig:fnn}  
\end{figure}

Feedforward neural networks (FNN) also referred to as multilayer perceptrons are directed layered neural networks with no internal feedback connections. In the mathematical sense, an FNN maps input vector $\mathbf{x}$ to output $y$, i.e., $f:\mathbf{x}\longrightarrow y$. An N-layered FNN is a composite function $y = f(\mathbf{x};\Gamma)=f_N(f_{N-1}(\cdots f_1(\mathbf{x})))$ mapping input vector $\mathbf{x}\in \mathbb{R}^m$ to a scalar output $y \in \mathbb{R}$. Here, $\Gamma$ represents the neural network parameters such as the weights and biases. Depth and width of the neural network are related to the number of layers in the neural network and number of neurons in the layers respectively. The layers in between the input and output layers for which the output does not show are collectively referred to as the \emph{hidden} layers. A 3-layered FNN accepting a two-dimensional input vector $\mathbf{x}\in \mathbb{R}^2$ approximating it to a scalar output $y \in \mathbb{R}$ is illustrated in Figure \ref{fig:fnn}.

Here, each node represents a neuron and each link between the nodes $i$ and $j$ are assigned a weight $w_{ij}$. The composite function of the 3-layered FNN is
\begin{equation}
    y = f(\mathbf{x};\Gamma) = f_3(f_2(f_1(\mathbf{x}))) \label{eq:fnn}
\end{equation}
In other words, the 3-layer FNN in Figure \ref{fig:fnn} is the directed acyclic graph equivalent of the composite function in equation (\ref{eq:fnn}). The subscript $n$ of $f_n$ indicates the layer number. The mapping in the first layer is 
\begin{equation}
    \mathbf{L}_1 = f_1(\mathbf{x}) = \gamma_1(\mathbf{W}_1\mathbf{x} + \mathbf{b}_1)
\end{equation}
where $\gamma_1(\circ)$ is the activation function, $\mathbf{b}_1$ is the bias vector, and $\mathbf{W}_1$ represents the weight matrix between the neurons in the first and second layers. Here, the weight matrix $\mathbf{W}_1$ is defined as the link weights between the neurons in the input and second layer
\begin{equation}
    \mathbf{W}_1 = \begin{bmatrix} w_{ab} & w_{db}\\w_{ae} & w_{de} \end{bmatrix}.
\end{equation}
Similarly, the second layer mapping can be represented as
\begin{equation}
    \mathbf{L}_2 = f_2(\mathbf{L}_1) = \gamma_2(\mathbf{W}_2\mathbf{L}_1 + \mathbf{b}_2)
\end{equation}
Finally, the output is
\begin{equation}
    y = f_3(\mathbf{L}_2) = \gamma_3(\mathbf{W}_3\mathbf{L}_2 + \mathbf{b}_3)
\end{equation}
The weight matrices in the second and final layers are
\begin{equation*}
    \mathbf{W}_2 = \begin{bmatrix} w_{bc} & w_{ec}\\w_{bf} & w_{ef} \end{bmatrix} \text{ and }
    \mathbf{W}_3 = \begin{bmatrix} w_{co} & w_{fo} \end{bmatrix}.
\end{equation*}
The neural network parameters $\Gamma = \{\mathbf{W}_1,\mathbf{W}_2,\mathbf{W}_3,\mathbf{b}_1,\mathbf{b}_2,\mathbf{b}_3 \}$ comprise the weight matrices and bias vectors across the layers. The objective of the training algorithm is to learn the optimal $\Gamma^*$ to get the target composite function $f^*$ from the available samples of $\mathbf{x}$.

\subsubsection{Convolutional Neural Networks}
\label{sec:cnn}
Convolutional networks or convolutional neural networks (CNNs) are a specialized type of feedforward neural network known for its spatial mapping capability. A CNN performs convolution operation in at least one of its layers. The \emph{feature extraction} capability of CNNs mimics the neural activity of the animal visual cortex \cite{CNNcortex}. The convolution operation in CNNs emulates the scene perception characteristic of the brain's visual cortex whereby they are sensitive to sub-regions of the perceived scene. Accordingly, CNNs have been widely used for computer vision problems \cite{googlenet_inception,alexnet,LeNet5,vgg16,squeezenet,cnn4vision,fastRCNN,CNNface,resnet}. The convolution is an efficient method of feature extraction that reduces the data dimension and consequently reduces the parameters of the network. Therefore, in contrast to its fully connected feedforward counterpart, CNNs are more efficient and easier to train. 

\begin{figure*}[h]
\centering
\includegraphics[width=4.7 in]{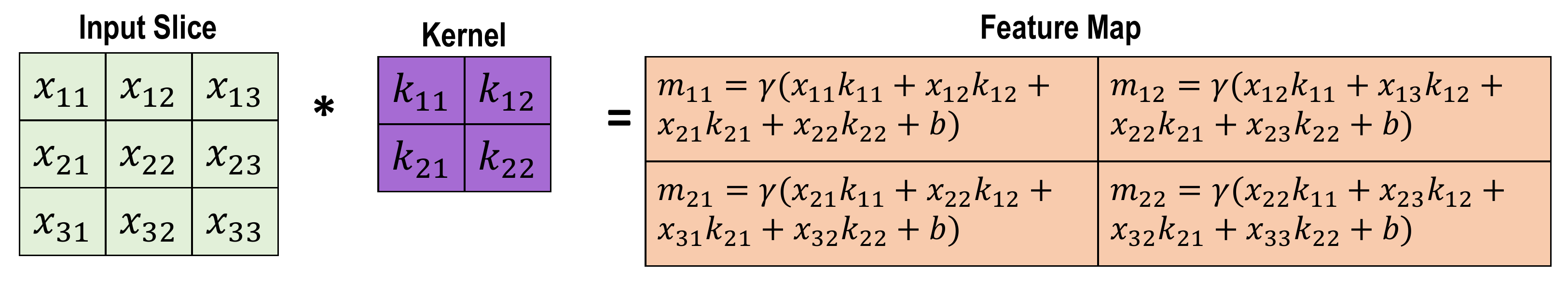}
\caption{Convolution of input slice with kernel}
\label{fig:cnn_conv}  
\end{figure*}

CNN architecture would often involve convolution, pooling, and output layers. The convolution layer convolve the input tensor $\mathbf{X}\in \mathbb{R}^{W\times H \times D}$ of width $W$, height $H$, and depth $D$ with the kernel (filter) $\mathbf{F}\in \mathbb{R}^{w\times h\times D}$ of width $w$, height $h$, and of the same depth as the input tensor to generate an output feature map $\mathbf{M}\in \mathbb{R}^{W_1\times H_1\times D_1}$. The dimension of the feature map is a function of the input as well as kernel dimensions, the number of kernels $N$, stride $S$, and the amount of zero padding $P$. Likewise, the feature map dimensions can be derived as $W_1 = \left(W-w+2P\right)/S + 1, \; H_1 = \left(H-h+2P\right)/S + 1,\; D_1 = N$. In other words, there will be as many feature maps as the number of kernels. Kernel refers to the set of weights and biases. The kernel operates on the input slice in a sliding window manner based on the stride - the number of steps with which to slide the kernel along with the input slice. Hence, each depth slice of the input is treated with the same kernel or in other words, shares the same weights and biases - \emph{parameter sharing}. A feature map illustration from a convolution operation on an input slice $\mathbf{x}$ by a kernel $\mathbf{f}$ is demonstrated in Figure \ref{fig:cnn_conv}. Here, $b$ represents the bias associated with the kernel slice and $\gamma\left(\circ\right)$ denotes a non-linear activation function.

\begin{figure}[h]
\centering
\hspace{-1 cm}
\includegraphics[width=2.1 in]{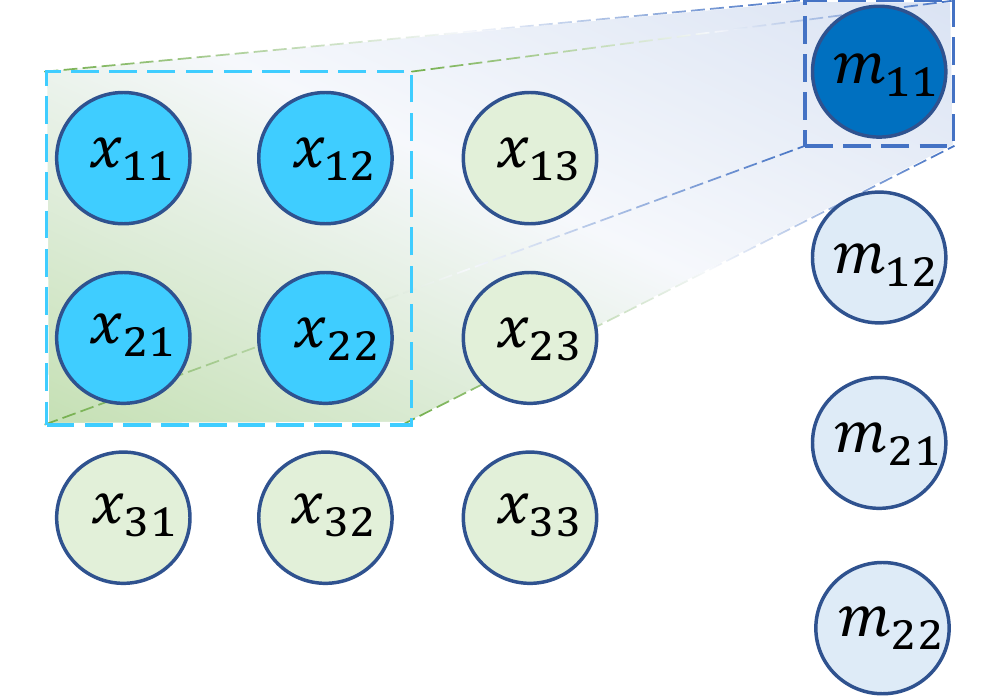}
\caption{Neural representation of convolution}
\label{fig:neural}  
\end{figure}

The resulting output from the convolution operation is referred to as the \emph{feature map}. Each element of the feature map can be visualized as the output of a neuron which focuses on a small region of the input - \emph{receptive field}. The neural depiction of the convolution interaction is shown in Figure \ref{fig:neural}. 

\begin{figure}[h]
\centering
\includegraphics[width=2.1 in]{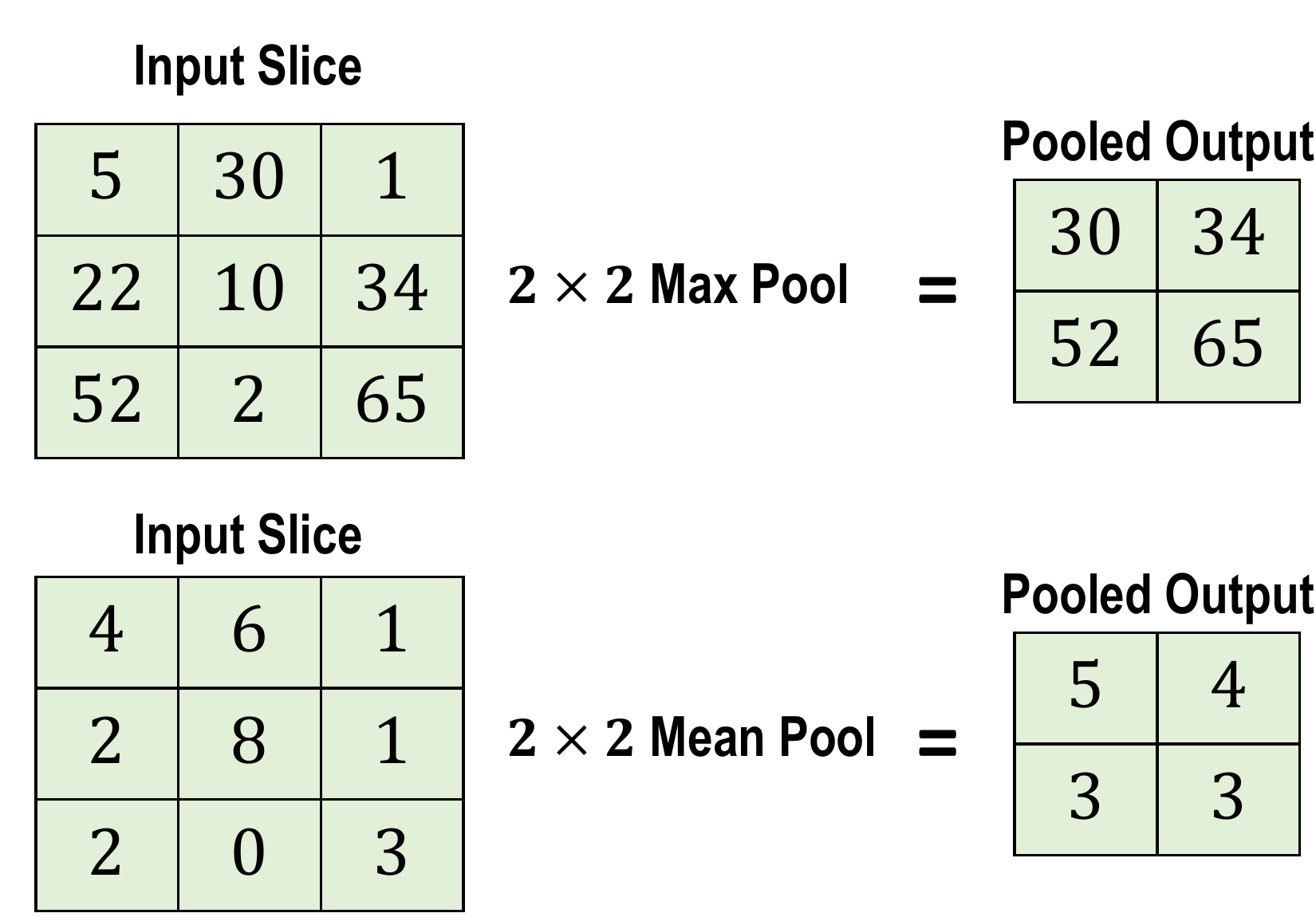}
\caption{Max and mean pooling on input slice with stride 1}
\label{fig:pool}  
\end{figure}

It is evident that each neuron in a layer is connected locally to the neurons in the adjacent layer - \emph{sparse connectivity}. Hence, each neuron is unaffected by variations outside of its receptive field while producing the strongest response for spatially local input pattern. The feature maps are propagated to subsequent layers until it reaches the output layer for a regression or classification task. \emph{Pooling} is a typical operation in CNN to significantly reduce the dimensionality. It operates on a subregion of the input to map it to a single summary statistic depending on the type of pooling operation - max, mean, $L_2$-norm, weighted average, etc. In this way, pooling downsamples its input. A typical pooling dimension is $2\times2$. Larger pooling dimensions might risk losing significant information. Figure \ref{fig:pool} shows max and mean pooling operations. 

A pooling layer of dimensions $W_p\times H_p$ upon operating over an input volume of size $W_1\times H_1\times D_1$ with a stride of $S_1$ will yield an output of volume $W_2 = \left( W_1-W_p\right)/S_1, \;H_2 = \left( H_1-H_p\right)/S_1, \; D_2 = D_1$. Pooling imparts invariance to translation, i.e., if the input to the pooling layer is shifted by a small amount, the pooled output will largely be unaffected \cite{Goodfellow-et-al-2016}.

The three essential characteristics of CNNs that contribute to the statistical efficiency and trainability are parameter sharing, sparse connectivity, and dimensionality reduction. CNNs have demonstrated superior performance in computer vision tasks such as image classification, object detection, semantic scene classification, etc. Accordingly, CNNs are increasingly used for UAS imagery and navigation applications \cite{uavapps}. Most notable CNN architectures are LeNet-5 \cite{LeNet5}, AlexNet \cite{alexnet}, VGG-16 \cite{vgg16}, ResNet \cite{resnet}, Inception \cite{googlenet_inception}, and SqueezeNet \cite{squeezenet}.

\subsubsection{Recurrent Neural Networks}
\label{sec:rnn}
Recurrent Neural Network (RNN) \cite{Rumelhart1986} is a specialized feedforward neural network tailored to capture temporal dependencies from sequential data by leveraging internal memory states and recurrent connections. Consequently, RNNs are well suited to solve sequential problems by exploiting the temporal correlation of data rendering them suitable for image captioning, video processing, speech recognition, and natural language processing applications. Moreover, unlike CNN and traditional feedforward neural networks, RNNs can handle variable-length input sequences with the same model. 

\begin{figure}[h]
\centering
\includegraphics[width=2 in]{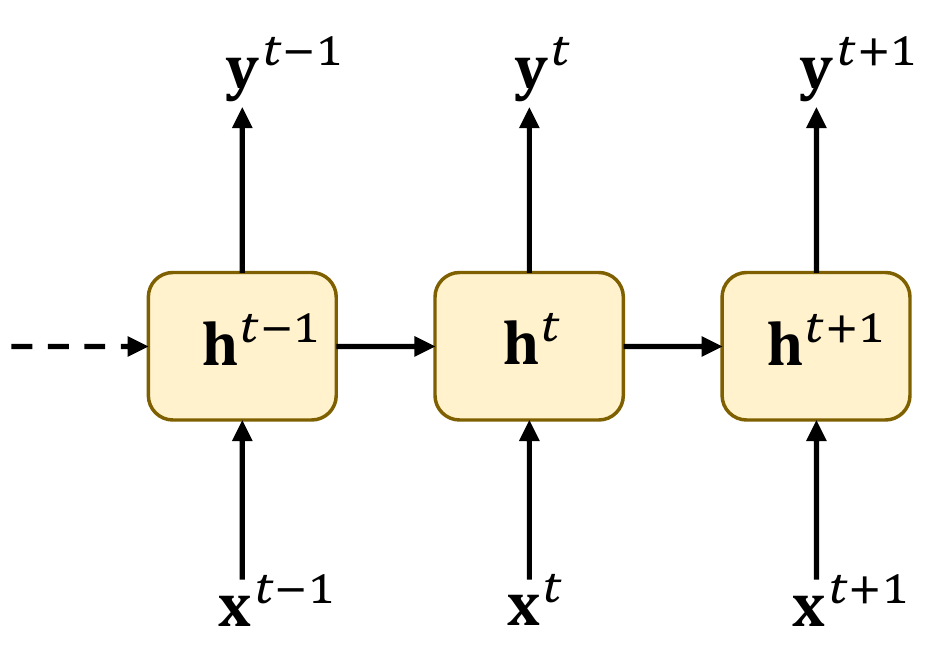}
\caption{Many-to-many RNN architecture}
\label{fig:mmrnn}  
\end{figure}

RNNs operate on input sequence vectors at varying time steps $\mathbf{x}^{t}$ and map it to output sequence vectors $\mathbf{y}^{t}$. The recurrence relation in an RNN parameterized by $\mathbf{\Gamma}$ can be expressed as 
\begin{equation}
    \mathbf{h}^t = \mathcal{F}\Big(\mathbf{h}^{t-1},\mathbf{x}^{t};\mathbf{\Gamma} \Big)
    \label{eq:recursive}
\end{equation}
where $\mathbf{h}^t$ represents the hidden state vector at time $t$. The recurrence relation represents a recursive dynamic system. By this comparison, RNN can be defined as \emph{a recursive dynamic system that is driven by an external signal, i.e, input sequence $\mathbf{x}^{t}$}. The equation (\ref{eq:recursive}) can be unfolded twice as
\begin{align}
  \mathbf{h}^t &= \mathcal{F}\Big(\mathbf{h}^{t-1},\mathbf{x}^{t};\mathbf{\Gamma} \Big)\\
  &= \mathcal{F}\Big(\mathcal{F}\Big(\mathbf{h}^{t-2},\mathbf{x}^{t-1};\mathbf{\Gamma} \Big),\mathbf{x}^{t};\mathbf{\Gamma} \Big)\\
  &= \mathcal{F}\Big(\mathcal{F}\Big(\mathcal{F}\Big(\mathbf{h}^{t-3},\mathbf{x}^{t-2};\mathbf{\Gamma} \Big),\mathbf{x}^{t-1};\mathbf{\Gamma} \Big),\mathbf{x}^{t};\mathbf{\Gamma} \Big)
\end{align}
The unfolded equations show how RNN processes the whole past sequences $\mathbf{x}^{t}, \mathbf{x}^{t-1},$ $\cdots, \mathbf{x}^{1}$ to produce the current hidden state $\mathbf{h}^{t}$. Another notable inference from the unfolded representation is the \emph{parameter sharing}. Unlike CNN, where the parameters of a spatial locality are shared, in an RNN, the parameters are shared across different positions in time. For this reason, RNN can operate on variable-length sequences allowing the model to learn and generalize well to inputs of varying forms. On the other hand, traditional feedforward network does not share parameters and have a specific parameter per input feature preventing it from generalizing to an input form not seen during training. At the same time, CNN share parameter across a small spatial location but would not generalize to variable-length inputs as well as an RNN. A simple many-to-many RNN architecture which maps multiple input sequences to multiple output sequences is shown in Figure \ref{fig:mmrnn}.

\begin{figure*}[h]
\centering
\includegraphics[width=5.8 in]{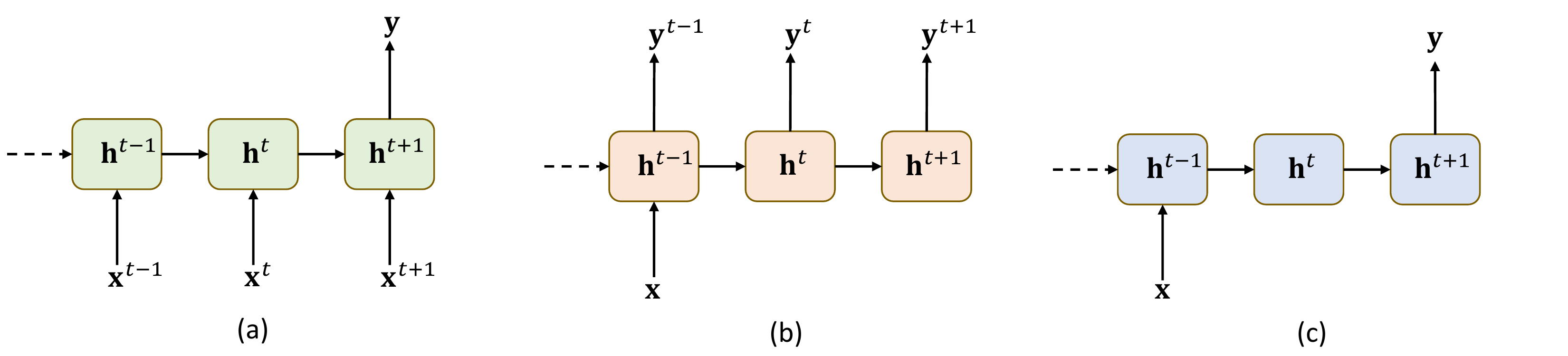}
\caption{RNN architectures. (a) Many-to-one, (b) One-to-many, and (c) One-to-one}
\label{fig:allrnn}  
\end{figure*}

For a simple representation, let us assume the RNN is parameterized by $\mathbf{\Gamma}$ and $\mathbf{\phi}$ with input-to-hidden, hidden-to-hidden, and hidden-to-output weight matrices being $\mathbf{W}_{ih}, \mathbf{W}_{hh},$ and $\mathbf{W}_{ho}$ respectively. The hidden state at time $t$ can be expressed as 
\begin{align}
   \mathbf{h}^t &= \mathcal{F}\Big(\mathbf{h}^{t-1},\mathbf{x}^{t};\mathbf{\Gamma} \Big)\\ 
   &= \gamma_h\Big(\mathbf{W}_{hh}\mathbf{h}^{t-1} + \mathbf{W}_{ih}\mathbf{x}^{t} + \mathbf{b}_h\Big).
\end{align}
where $\gamma_h(\circ)$ is the activation function of the hidden unit and $\mathbf{b}_h$ is the bias vector. The output at time $t$ can be obtained as a function of the hidden state at time $t$,
\begin{align}
   \mathbf{y}^t &= \mathcal{G}\Big(\mathbf{h}^{t};\mathbf{\phi} \Big)\\
   &= \gamma_o\Big(\mathbf{W}_{ho}\mathbf{h}^t + \mathbf{b}_o\Big)
\end{align}
where $\gamma_o(\circ)$ is the activation function of the output unit and $\mathbf{b}_o$ is the bias vector. RNN could take different forms such as many-to-one, one-to-many, and one-to-one as illustrated in Figure \ref{fig:allrnn}.

The RNN architectures discussed here captures only hidden states from the past. Some applications would also require future states in addition to past. This can be accomplished by a bidirectional RNN \cite{biRNN}. In simple words, bidirectional RNN combines an RNN that depends on past states (\emph{i.e.,} from $\mathbf{h}^{1}, \mathbf{h}^{2}, \mathbf{h}^{3}, \cdots, \mathbf{h}^{t}$) with that of an RNN which looks at future states (\emph{i.e.,} from $\mathbf{h}^{t}, \mathbf{h}^{t-1}, \mathbf{h}^{t-2}, \cdots, \mathbf{h}^{1}$). In RF applications, RNNs may be used with time series data for spectrum forecasting, spectrum usage pattern analysis, anomaly detection, among others.

\subsubsection{Generative Adversarial Networks (GANs)}
\begin{figure}[h]
    \centering
    \includegraphics{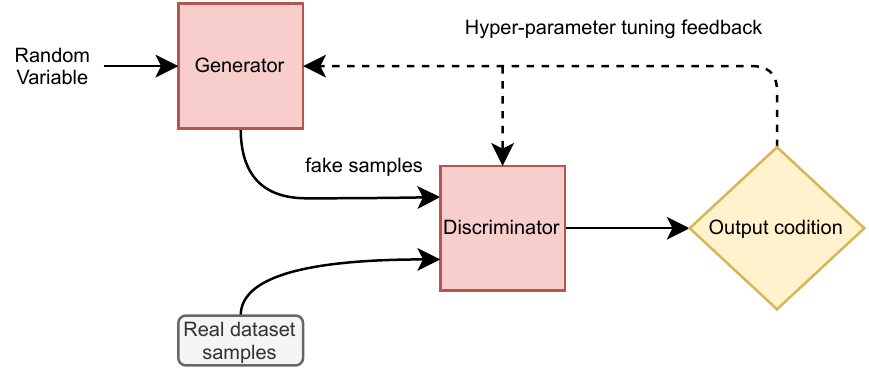}
    \caption{Simple Generative Adversarial Networks (GANs) structure}
    \label{fig:gans_structure}
\end{figure}

GANs is a machine learning framework that consists of two neural networks that compete against each other \cite{goodfellow2014generative}. The two networks are called the Generative network (Generator \emph{G}) and Discriminative network (Discriminator \emph{D}) as shown in Figure \ref{fig:gans_structure}. The generator \emph{G} generates samples from the model distribution and learns to deceive the discriminator \emph{D}, while the discriminator learns to distinguish between samples from dataset and samples from generator. The generative model generates samples by passing random noise through an FNN, and the discriminative model is also built using an FNN and outputs a scalar $D(x)$ that represents the probability that the samples are from the dataset. Generative model \emph{G} is represented by $G(z;\Gamma_g)$, where $\theta_g$ is the FNN paremeters and $z$ is the input noise variable with probability distribution $p_z(z)$. The discriminator model \emph{D} is represented as $D(x;\theta_d)$ where $\theta_d$ is the FNN parameters and $x$ is the input data samples. \emph{D} is trained to maximize the probability of assigning correct label to the samples from dataset and \emph{G} is trained to minimize $\log (1-D(G(z)))$. The value function $V(G,D)$ of this minimax game of \emph{D} and \emph{G} is given by,

\begin{multline}
    \mathop{\mathrm{min}}_{G} \mathop{\mathrm{max}}_{D} V(D,G) = \mathbb{E}_{x\sim p_{data}(x)}[\log D(x)]\\
    +\mathbb{E}_{z\sim p_{data}(z)}[\log (1-D(G(z)))]
\end{multline}

In each training epoch, the discriminator is trained first for a fixed number of steps by inserting real and fake data samples, and update the discriminator by ascending the stochastic gradients by keeping the generator fixed. Once the discriminator is trained for a fixed number of steps, the generator is updated by descending its stochastic gradient while keeping its discriminator fixed and inserting fake data with fake labels to deceive the discriminator. 


\subsection{CNN based RF fingerprinting}
\subsubsection{ORACLE}
The authors propose a CNN framework for RF fingerprinting called ORACLE (Optimized Radio clAssification through Convolutional neuraL nEtworks) \cite{sankhe2019oracle}. This work provides one of the most extensive evaluations where they demonstrate up to $99\%$ classification accuracy on more than 100 consumer-of-the-shelf (COTS) WiFi devices. They also demonstrate similar results on 16 bit-similar USRP X310 SDRs. Other key contributions of this work include the study of hardware-driven features occurring in the transmit chain that causes IQ sample variation. They study both static and dynamic channel environment. In the case of the dynamic channel, they explore how feedback-driven transmitter-side modifications that use channel estimation at the receiver can increase the differentiability for the CNN classifier. Essentially, introducing perturbations/imperfections on the transmitter-side to aid classification while minimizing the impact on bit error rates. 

Specifically, in the context of studying the effects of hardware driven RF impairments, the authors focus on IQ imbalance and DC offset. They use MATLAB Communications System Toolbox to generate IEEE 802.11a standard compliant packets. The transmitter in this case was a USRP X310 and the receiver was a USRP B210. They also use an external database which consisted of raw IQ collected from 140 devices which included phones, tablets, laptops, and drones belonging to 122 manufacturers.

For the case of static channel, the authors used the following architecture;
\begin{itemize}
    \item \textbf{Input}: Raw IQ with length $128$. This was formatted into two-dimensional real value tensor of size $2 \times 128$
    \item \textbf{Network}: Two convolutional layers and two fully connected layers each with 256 and 80 neurons.
    \item \textbf{Kernels}: 1st layer consists of $50$ $1 \times 7$ filters, second layer consists $50$ $2 \times 7$ filters \item \textbf{Activation Function}: Each convolution layer is followed by ReLU activation
    \item \textbf{Output}: A softmax is used in the last layer for classification. 
\end{itemize}

This architecture provided a median classification accuracy of $99\%$ when up to 100 different devices were used and the performance dropped slightly to $96\%$ for $140$ devices. For the dataset collected using 16 X310 radios, the accuracy was close to $98.6\%$. 

The authors clearly highlight the challenge put forth by dynamic channels and propose introducing controlled impairments to the transmitter as a solution to alleviate it. While this may work in a setting where one can make such impairments to the transmit chain, in many commercial and tactical applications this is not an option at the physical layer. It can be further argued at that point one is not exactly exploiting the "unique" RF fingerprint of the device but rather assigning the device an artificial tag/id/fingerprint. This is certainly an interesting area of research that may have its application and advantages but will also have limitations and disadvantages.

\subsubsection{Unmanned Aerial Vehicles With Non-Standard Transmitter Waveforms}
The authors propose a multi-classifier-based RF fingerprinting for UAVs \cite{UAV_multiClass_2020}. This work aims to combine outputs of multiple deep neural networks trained on a different portion of the training set. The authors create a dataset by collecting signals from 7 identical DJI M100 UAVs using an USRP X310 equipped with a UBX 160 daughterboard in an RF anechoic chamber. The IQ signals are captured by flying the UAVs at a distance of 6, 9, 12, and 15 feet from the receiver. They collect four bursts of ~2 seconds, each burst containing $\sim$140 data sequences (examples) for all the UAVs at the four different distances individually. The authors use 1D modified versions of AlexNet (AlexNet1D) and ResNet50 (ResNet1D) neural network architectures for classification. AlexNet1D is a forward CNN with five blocks (consisting of two 1D convolutional layers with 128 filters of sizes 7 and 5 respectively, followed by a MaxPooling layer) stacked on top of 2 fully connected layers of sizes 256 and 128 respectively.

This work is the first to show the effect of aerial hovering of UAVs on the accuracy of DL-based RF fingerprinting. When the network is trained with all 4 bursts of the UAVs dataset, the network performs well in identifying the UAVs. But when the network is trained using the first three bursts and tested on burst 4, both the network architectures perform with an accuracy of ~$50\%$. This drop in accuracy shows the effect of continuous channel variation and minute UAV movements when hovering. To overcome this effect, the authors propose a multi-classifier scheme in which the burst signals from the dataset are partitioned to form non-overlapping sets and each of these partitions is used as training sets for identical but independent AlexNet1D Neural Networks. The outputs of the neural networks are then combined using a two-level score-based aggregation method. They also propose an algorithm for determining the number of neural networks to be used in the multi-classifier scheme. To evaluate the proposed method, the authors choose to use 12 neural networks. The proposed multi-classifier technique improves the classification accuracy from $50\%$ when using a single classifier to $91\%$ when the network is trained using the first three bursts and tested using the fourth burst.

The authors also propose a Data Augmentation (DA) scheme for training individual neural networks in the multi-classifier technique. DA is the method of expanding the training dataset by modifying the original samples in a proper manner \cite{shorten2019survey}, \cite{MoreIsBetter2020soltani}. In this work, DA is performed by normalizing the training batch according to the mean and standard deviation of the whole dataset. The normalized dataset is then passed through the DA block where it is convolved with a block of multi-tap complex FIR filters. The use of DA improved the accuracy of the multi-classifier technique by up to $95\%$. This work also proposes a method for detecting new UAVs (UAVs not included in the training dataset). Using the proposed multi-classifier with a data augmentation scheme, the authors show an accuracy of $99\%$ in detecting new UAVs. The authors clearly state that the improvement in the accuracy when using the multi-classifier approach comes with no increase in model size compared to single ResNet1D architecture but at the cost of a longer testing/training process. However, one can claim that the data capture in an anechoic chamber eliminates the rich multipath propagation effects as in the real-world settings.

\subsubsection{SEI using the bispectrum}
In \cite{ding2018SEI_CNN}, the authors propose a deep learning-based SEI using the bispectrum of the received signal as the feature. The bispectrum is estimated by calculating the third-order cumulant of the RF signal. Further, the bispectrum dimensions are reduced (bispectrum compression) using the projection method in \cite{Yong2009neurocomputing}. The reduced bispectrum is then fed into a CNN consisting of three convolution layers ($30$ kernels of size $3\times3$), a fully connected layer with $128$ neurons, and a final softmax layer that maps the outputs to their respective classes.

Signals are collected from five USRPs including, one E310, three B210, and one N210, and the authors show that the proposed method has an accuracy of $75\%$ in identifying the five USRPs. They also collect signals from ten emitters modeled using a memory polynomial model that consists of multiple delays and nonlinear functions \cite{sappal2015memoryPolynomial}. The proposed model has an accuracy of $85\%$ in identifying ten modeled emitters and $87\%$ in identifying five modeled emitters.

\subsubsection{Differential Constellation Trace Figure (DCTF)}
The authors of \cite{peng2020DCTF} propose the use of DCTF to extract RF fingerprint features and use a CNN to identify different devices using the DCTF features. The DCTF is a 2D representation of the differential relation of the time-series signal. DCTF-based feature extraction was first proposed in \cite{peng2016zigbee_dctf} where they used a minimum distance classifier to achieve an accuracy of $90\%$ in identifying 16 CC2530 ZigBee modules at SNR$\geq15$ dB. In this work, the authors aim to use CNN as a classifier to improve classification accuracy. The DCTF figure is highly influenced by the hardware imperfections that are related to the RF fingerprint features. These images are classified using a CNN to identify the devices. To evaluate the performance of the DCTF-CNN, the authors capture signals from 54 Texas Instruments CC2530 ZigBee modules using a USRP. The DCTF is computed for each of the signals and classified using a network consisting of three convolutional layers and one fully connected layer. The three convolutional layers are of sizes 16, 32, and 64, respectively, with a kernel size of $3\times3$. A $2\times2$ max pooling is applied to each of the outputs of the convolutional layers. 

The performance of the DCTF-CNN method is evaluated for different DCTF image quality and SNR. DCTF image quality depends on the size of the DCTF size. Lower size DCTF images perform poorly because of blurring of features, whereas larger size DCTF images have better performance but with the drawback of requiring more samples and higher complexity. In this work, the best performance for the designed CNN is achieved by using a DCTF image of size 65x65. Using the fixed DCTF image size, the authors further investigate the effect of SNR on performance. The DCTF-CNN achieves a classification accuracy of $93.8\%$ at SNR of 15dB and $99.1\%$ at SNR 30dB.

\subsubsection{RF signal spectrum}
In \cite{zong2020cnn}, the authors propose the use of a CNN to identify the devices from the RF signal spectrum. A dataset consisting of 10,000 signals from each of the five transmitters at an SNR of 20 dB generated by Monte Carlo experiments with random AWGN and multipath channels is used. The RF signals are processed by an STFT to convert the time domain signals to time-frequency domain thereby generating the RF signal spectrum. RF signal spectrum reflects the characteristics of the signal in the frequency domain and the change of the frequency domain of the signal over time. The RF signal spectrum is then fed into the CNN to classify the signal.

\begin{figure}[h]
    \centering
    \includegraphics{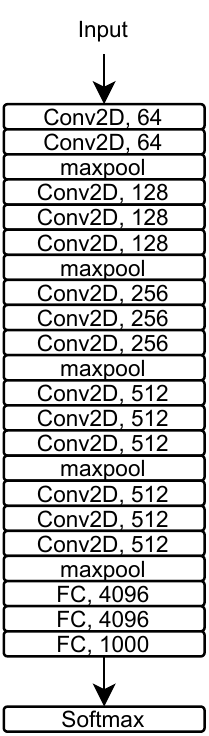}
    \caption{VGG-16 Network}
    \label{fig:vgg-16}
\end{figure}

The authors use a modified version of the VGG-16 model to classify the signals. VGG-16 network model consists of thirteen convolution layers with a kernel of $3\times3$ and two fully connected layers interlaced with five maxpool layers, as shown in Figure \ref{fig:vgg-16}. The output of the final layer is fed to a softmax layer to generate transmitter class tags distribution. The VGG-16 is modified by adding a Batch Normalization (BN) operation after each convolutional layer and a random dropout layer after the first two fully connected layers. BN helps in speeding up the model's convergence during training, and the dropout layer discards random neurons leading to a more sparse feature map thereby helping in reducing overfitting. The network is trained with 1000 signals from each of the five transmitters using an Adam optimizer to minimize the loss. The proposed method achieves an accuracy of $99.7\%$ in identifying five devices.

\subsubsection{A Massive Experimental Study}
\label{sec:a massive ecperimental study}
A large-scale RF fingerprinting study on a DARPA dataset - 400 GB of WiFi and Automatic Dependent Surveillance-Broadcast (ADS-B) waveforms from 10,000 devices - is presented in \cite{jian2020dl_rff_massive}. This is the first large-scale study which elaborates the effect of device population, burst type, environmental, and channel effects on CNN-based RF fingerprinting architectures. The authors present two architectures: 1) A baseline model inspired by AlexNet comprising five stacks followed by four fully connected layers. Each stack is composed of two convolution layers (128 kernels each with kernel sizes $1\times 7$ and $1\times5$ respectively) and a max pooling layer, 2) A ResNet-50-1D which is a modified ResNet-50 architecture such that it can accommodate one-dimensional time series IQ samples. The WiFi dataset include emissions from 5117 devices with 166 transmissions on average from each device while the ADS-B dataset contains 5000 emitters of 76 average emissions. The authors preprocess the WiFi dataset with band filtering and partial equalization and adopt a sliding window approach to transform signals in both datasets to a fixed form suitable for the CNNs. 

The authors conduct an extensive RF fingerprinting study in parts by forming multiple learning tasks for the CNN classifiers. In order to ease the readers into these 22 learning tasks, we succinctly list the broad task categories below:
\begin{enumerate}
    \item Task 1 - network performance with scaling device population grouped into four categories (A to D).
    \item Task 1M - similar to Task 1 but under multiburst setting where each device may emit multiple transmissions for joint classification. Here again there are four subtasks (A to D).
    \item Task 2 - effect of training set size on classifier performance. This task has three subtasks (A to C).
    \item Task 3 - effect of channel by collecting captures under varying time frames and environmental conditions (indoor vs outdoor). This task has four subtasks (A to D).
    \item Task 4 - assess the effect of SNR on classification accuracy by four subtasks (A to D).
    \item Task 5 - with 19 bitwise identical emitters.
\end{enumerate}

The dataset is grouped into different subsets to suit the aforementioned tasks. With the above task assessments, the authors reported that both baseline and ResNet-50-1D models scale gracefully
on Task 1. The multiburst predictions, i.e., Task 1M along with Task 5 demonstrated significantly higher accuracy. Task 2 evaluation indicated improvement in model accuracy with the inclusion of more transmissions in the training set. Finally, the Tasks 3 and 4 exhibited that the predictions were affected by environmental and channel conditions. The authors state that the ADS-B emitter classification manifested as a simpler classification problem in contrast to WiFi in light of the higher accuracy. Finally,
the baseline model which is the modified AlexNet demonstrated superior performance in comparison to ResNet-50-1D in several of the learning tasks, indicating deeper is not always better. These results are also discussed in \cite{shawabka2020exposing} where evaluation on a custom dataset collected by the authors in a laboratory setting is also presented. This custom dataset is a 7 TB dataset comprising emissions from 20 USRP radios. We elaborate on this dataset in section \ref{sec:dataset}.
\subsubsection{Trust in 5G Open RANs}
Reus-Muns et al. in \cite{reusmuns2019trust} propose the use of CNNs augmented with triplet loss to detect specific emitters through RF fingerprinting. The authors aim to combat the adversarial impact of the wireless channel by using a neural network with a triplet-loss function. A dataset consisting of signals from Basestations that emit standards-compliant WiFi, LTE, and 5G New Radio (NR) waveforms is used to evaluate the proposed network. The dataset is described in detail in Section.\ref{sec:dataset_POWDER}. The dataset is used to train two CNN classifiers: baseline CNN and Triplet network. The baseline CNN architecture consists of four convolutional layers (with 40 filters of size $1\times7$, $1\times5$, $2\times7$, and $2\times5$, respectively) followed by three fully connected layers and a final softmax classifier layer. The Triplet network architecture is similar to the baseline CNN except a triplet loss function is employed. The triplet loss \cite{Schroff_2015_CVPR} is designed to enforce class separation into embedding space and is trained on a series of triples - anchor, positive, and negative. The triplet loss function aims to train the neural network to maximize the separation between the anchor and the negative labels while minimizing the distance between the anchor and the positive classes.

The proposed baseline and triplet loss CNN are trained and tested with the WiFi transmissions from the dataset and the overall classification accuracy is $92.92\%$ and $99.98\%$, respectively. Next, the authors propose three step algorithm that returns a quantitative measure of trust in a Base Station (BS). This approach assigns a trust category based on the softmax probability range of the chosen class. For softmax range $\leq80\%$ the device is assigned \emph{No Trust}, for the range $[80\%,\;99\%]$ the device is tagged as \emph{Partial Trust}, and for $\geq 99\%$ the device will fall under \emph{Trusted} category.
 
\subsubsection{Dilated Causal Convolutional Model}
The authors propose an augmented dilated causal convolution (ADCC) network that combines a stack of dilated causal convolution layers with traditional convolutional layers to classify wireless devices based on their RF fingerprints \cite{Robinson2020ADCC}. In this work, the authors train and evaluate the proposed model on transmissions from up to 10,000 devices consisting of WiFi (IEEE 802.11a and 802.11g) and ADS-B signals. The authors use the data provided by Radio-Frequency Machine Learning Systems (RFMLS) research program \cite{RFMLS}. It consists of 103 million transmissions from over 53,000 WiFi devices and 3.5 million ADS-B transmissions from over 10,000 devices.

\begin{figure}[h]
    \centering
    \includegraphics[width=0.48\textwidth]{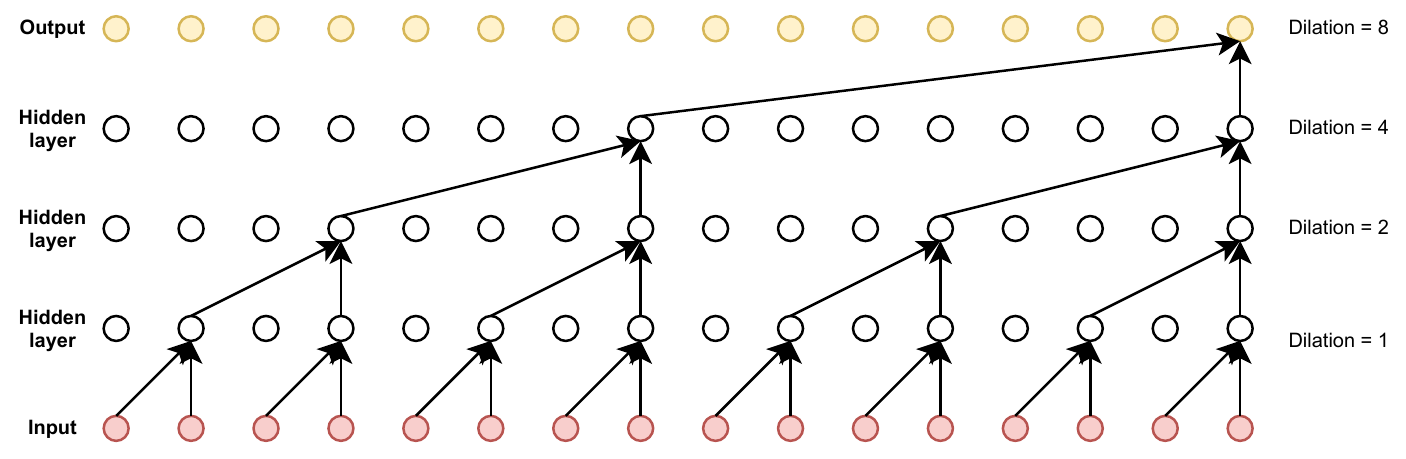}
    \caption{Dilated Causal Convolutional Network}
    \label{fig:DCCN}
\end{figure}

A dilated convolution is a type of convolution where the filter is applied over an area larger than its length by skipping input values with a certain step as shown in Figure \ref{fig:DCCN} \cite{oord2016wavenet}. 

The proposed ADCC model consists of two main components, the residual blocks, and the traditional convolution, and pooling blocks. The model consisted of eight residual blocks in series, each made up of a gated convolution operation that is causal and dilated (GDCC) followed by a causal convolution layer with a kernel size of one. Before the first residual block, the input is passed through a DCC layer with dilation rate of one. Skip connections from each of the eight residual blocks are summed and used as the input for the traditional convolution and pooling blocks, consisting of three stacks of two convolution layers and a pooling layer each. The first 1600 IQ values are processed by the ADCC model to generate 2500 features. To extract additional features from ID-specific information about the device, the authors propose to extract 2500 features from twenty subsequences of size 200 IQ values uniformly distributed throughout the rest of the signal. Each of the twenty subsequences is processed by two blocks with causal convolution and pooling layer, individually. The feature map from each of the twenty processes is then input to an average pooling resulting in 2500 features. The 2500 features from the ADCC block, and the 2500 features extracted from the twenty subsequences are concatenated together and passed as the input to the dense classification layer.

The authors evaluate the proposed method by conducting the learning tasks Task 1 through Task 4 as in section \ref{sec:a massive ecperimental study}. It was noted that for Task 1 the performance scaled linearly in  the logarithm of the device population.  The multiburst accuracy was shown to be better than single burst accuracy implying performance gains with coherent processing. Similar to the \cite{jian2020dl_rff_massive}, the accuracy of ADS-B device fingerprinting was higher in contrast to WiFi which the authors state could be due to the open air propagation of ADS-B. The Task 2 experiments further revealed only 2\% drop in accuracy when training size is reduced from 501 to 313 suggesting the network efficiency with small training size. The authors note drastically reduced performance under Task 3 evaluations when the channel differs between the training and validation sets implying the sensitivity of fingerprint features to the propagation effects. The evaluation in Task 4 exhibited lower accuracy when the training SNR was higher than the validation SNR, and higher accuracy when validation SNR was higher than the training SNR.  

In a more recent work \cite{kudzebaRiftNet}, the authors adopt a multi-burst approach towards improving the fingerprinting accuracy for large-scale fingerprinting involving greater population sizes. Specifically, the multi-burst processing utilizes multiple bursts of the signal from the same but unknown device (i.e., sharing same label) to drive the noise level down. The inference is performed on multiple bursts and the class probability vectors from the inference on each bursts are combined to derive a final class prediction. The authors report a prediction accuracy in upwards of 95\% across different signal types (WiFi and ADS-B).

To conclude this discussion on RF fingerprinting using CNNs, we have enlisted the reviewed works in a tabular form in Table \ref{tab:CNN_label} for easy reference.
\begin{table*}[h!]
    \centering
    \begin{tabularx}{0.9\textwidth}{>{\raggedright\arraybackslash}X >{\raggedright\arraybackslash}X >{\raggedright\arraybackslash}X >{\raggedright\arraybackslash}X  }
         \hline
         Work & RFF feature & RF emitters & Performance \\
        \hline
        \hline
        \emph{Sankhe et al.}\cite{sankhe2019oracle}& IQ imbalance \& DC offset & 140 devices (phones, tablets, laptops, \& drones) & $99\%$ up to 100 devices, $96\%$ up to 140 devices \& $98.6\%$ for dataset\cite{oracle_dataset} \\\\
        \emph{Soltani et al.}\cite{UAV_multiClass_2020}& Multiple data bursts & 7 DJI M100 UAVs & up to $99\%$  \\\\
        \emph{Ding et al.}\cite{ding2018SEI_CNN}& Bispectrum & 1 E310, 3 B210s, \& 1 N210 & up to $87\%$  \\\\
        \emph{Peng et al.}\cite{peng2020DCTF}& DCTF &54 Texas Instruments CC2530 ZigBee modules & $93.8\%$ at 15dB SNR and $99.1\%$ at 30dB SNR\\\\
        \emph{Zong et al.}\cite{zong2020cnn}& RF signal spectrum & 5 simulated transmitters  & $99.7\%$  \\\\
        \emph{Jian et al.} \cite{jian2020dl_rff_massive}& Time-domain RF Signal &5117 WiFi devices and 5000 ADS-B devices & Per-transmission ADS-B accuracy of $91.9\%,\;76.8\%,\;92.5\%$ with 100 devices for Task 1D, 2C, and 4F, respectively.\\\\
        \emph{Amani Al-Shawabka et al.}\cite{shawabka2020exposing}& Time-domain RF Signal & 20 National Instruments SDR (12 NI N210 and 8 NI X310) & Training and Testing on the same day $\geq87.41\%$\\\\
        \emph{Guillem Reus-Muns et al.}\cite{reusmuns2019trust}& Time-domain RF Signal & 4 BSs in the POWDER Platform\cite{powder_platform} & $99.98\%$ for 10  slices using majority voting\\
       
        \emph{Josh et al.}\cite{Robinson2020ADCC}& Time-domain RF Signal & 53k WiFi and 10k ADS-B devices  & Top-five accuracy of $97\%,\;95\%,\;99\%,\;98\%$ with 100 devices for Task 1D, 2C, 3E, and 4F, respectively.\\
        \hline
    \end{tabularx}
    \caption{CNN architectures for RF fingerprinting}
    \label{tab:CNN_label}
\end{table*}

\subsection{Generative Adversarial Networks}

\subsubsection{Classification based on Auxiliary Classifier Wasserstein GANs}
The authors propose a RF-based UAV classification system based on Auxiliary Classifier Wasserstein Generative Adversarial Networks (AC-WGAN) in \cite{AC-WGANs2018zhao}. In this work, the authors collect wireless data from four different types of UAVs (including Phantom, Mi, Hubsan, and Xiro) using Agilent (DSO9404A) oscilloscope with antenna for indoor environment and USRP N210 with CBX daughterboard for an outdoor environment. The authors show the proposed system achieves an accuracy of $95\%$ for recognizing UAVs in an indoor environment. 

In this work, the authors improve the discriminative network of the auxiliary classifier generative adversarial nets (AC-GAN) proposed in \cite{odena2017conditional} to modify it to classify wireless signals collected from UAVs and also improve the AC-GAN model following the Wasserstein GAN (WGAN) \cite{arjovsky2017wasserstein} model to make the proposed model more stable during training. RMSProp is chosen as the loss function instead of Adam due to its better performance in nonstationary problems \cite{gulrajani2017improved}. During training, samples (training samples) are fed to the generator and the discriminator networks to update the negative critic loss by ascending its stochastic gradient. Then, the testing samples are classified by the discriminator according to the value of negative critic loss.

The authors capture the wireless signals from the UAVs and apply a bandpass filter to get the wireless signal in the 2.4-2.5 GHz band, after which the start point of the signal is detected, and the amplitude envelope is extracted. To reduce the dimensionality of the UAV's wireless signals, the authors propose to use a modified PCA.
Using the signals captured indoors from the 4 UAVs and WiFi signal, the authors test the proposed model and show that the accuracy of classification of different UAVs is greater than $95\%$ at SNR of 5 dB. They also show that the proposed classification using AC-WGANs with PCA performs better than the standard SVM and AC-GAN models and are also suited for real-time classification over long distances in the range 10 m to  400 m.

\subsubsection{GANs with Adversarial learning}
In \cite{roy2019GANSs}, an adversarial learning technique for identifying RF transmitters is implemented using generative adversarial nets (GANs). The authors also implement a CNN and DNN based classifier that exploits the IQ imbalance present in the received signal to learn the unique fingerprint features for classifying the devices. The dataset used in this work consists of QPSK modulation signals from eight USRP B210s received using an RTL-SDR and are considered as signals from trusted transmitters. With the help of an adversary, the generator model generates random signals with noise and is passed to the discriminative model as input. The discriminative models of the GANs take input from both the generator model and trusted transmitters and improve the random signal to imitate the real data by giving feedback to the generator model for tuning the hyperparameters. The generator model network consists of two fully connected layers with 512 and 1024 nodes, and the discriminative model network is made of three fully connected layers with 1024, 512, and 2 nodes with dropout layers between the first two fully connected layers. The GANs is modeled to identify if the signal is from a trusted transmitter. The proposed GANs model identifies the 8 trusted transmitters with an accuracy of $99.9\%$.

\begin{figure}[h]
    \centering
    \begin{subfigure}{0.5\linewidth}
         \centering
         \includegraphics[width=0.5\columnwidth]{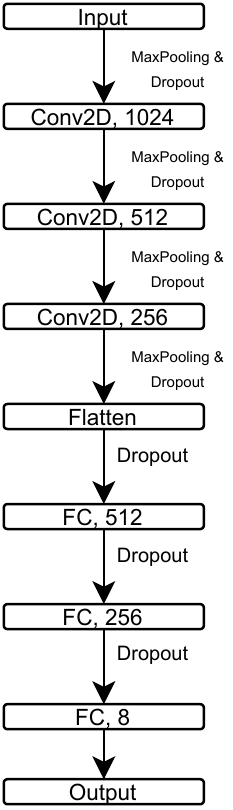}
         \caption{CNN}
         \label{fig:gans_cnn}
    \end{subfigure}
    \hfill
    \begin{subfigure}{0.45\linewidth}
         \centering
         \includegraphics[width=0.5\columnwidth]{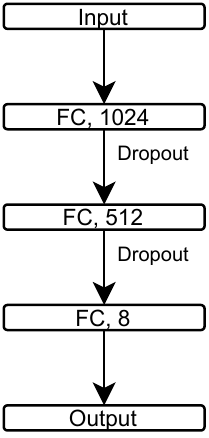}
         \caption{DNN}
         \label{fig:gans_dnn}
    \end{subfigure}
    \caption{Proposed CNN and DNN networks to classify signals}
    \label{fig:gans_with_dversarial}
\end{figure}

To classify the signals to identify the transmitters, the authors design and implement a CNN and a DNN. The CNN as shown in Figure \ref{fig:gans_cnn} has three Conv2D layers of size 1024, 512, and 256 filters with a kernel of size $2\times3$, followed by three fully connected layers of 512, 256, and 8 nodes. A MaxPooling2D layer of size $2\times2$ is applied after each of the three convolution layers and a dropout layer is applied after each of the convolution and fully connected layers. The DNN as shown in Figure \ref{fig:gans_dnn} has three fully connected layers with 1024, 512, and 8 nodes with dropout layers between the first two fully connected layers. The proposed CNN and DNN have a classification accuracy of $89.07\%$ and $97.21\%$, respectively, for classifying four devices and $81.59\%$ and $96.6\%$, respectively, for eight USRP devices. The classification accuracy of DNN is improved to $99.9\%$ by using the proposed GANs model to distinguish trusted transmitters from fake ones before classifying them.

\subsection{Probabilistic Neural Network}
\subsubsection{Energy spectrum based approach}
The authors in \cite{kose_energy_spectrum} propose a transient-based RF fingerprinting approach for extracting the unique characteristics of the wireless device from part of the energy spectrum of the transient signal. The work aims at reducing the computational complexity of feature extraction compared to techniques based on spectral fingerprinting. The proposed method is evaluated using data collected from eight IEEE 802.11b WiFi transceivers.

In this work, the instantaneous amplitudes of the captured signals are used to detect the transient. The starting point of the transient is estimated by using a Bayesian ramp change detector\cite{ureten2005bayesian}, and the endpoint is estimated by using a sliding window to calculate the average instantaneous amplitude of the samples. The first peak before the next steady state signal starts is chosen as the end point. Then the energy spectrum is obtained using discrete Fourier transform (DFT). By examining the frequency domain energy spectrum of these transients, the authors deduce that most of the information is concentrated in low-frequency components of the spectrum. Hinging on these observations, the authors propose that the number of energy spectral coefficients $(K)$ that carry the characteristic information can be calculated as,
\begin{equation}
    K = \left[ \frac{W}{\Delta f}  \right] \label{eq:transient_features}
\end{equation}
where $[\cdot]$ denotes integer part of $K$, $W$ is the transmission bandwidth, and $\Delta f$ is frequency resolution of the DFT given by
\begin{equation}
    \Delta f = \frac{1}{T_d} = \frac{1}{NT_s} = \frac{f_s}{N} \label{eq:transient_dft_res}
\end{equation}
where $N$ is the transient duration (in samples), $T_d$ is average transient duration in seconds, $T_s$ is sampling period, and $f_s$ is sampling frequency. Lastly, the classification is carried out by using a probabilistic neural network (PNN) classifier. A PNN is a feedforward neural network that estimates the probability density function (PDF) of each class using the Parzen window \cite{DudaRichardO2012Pcse}. Then, using the PDF of each class for the given input, the class with the highest posterior probability is estimated using Bayes' rule.

The classification performance of the proposed technique was carried out using a dataset collected from eight different IEEE 802.11b WiFi devices with 100 transmissions from each device. An average transient duration was calculated for the data in the training set for a sampling rate of $5$ GSps to determine the spectral feature length using equations (\ref{eq:transient_features}) and (\ref{eq:transient_dft_res}). The proposed method has a classification accuracy of $90\%$ and $97.91\%$ at $0$ dB and $25$ dB, respectively.

\subsubsection{Effect of Sampling Rate on Transient-based fingerprinting}
The authors in \cite{sampling_rate2017kose} investigate the effect of sampling rate on the classification performance of a transient-based RF fingerprinting method. A Bayesian ramp detector is employed to detect turn-on transient and amplitude features (instantaneous amplitude responses), and its dimensionality reduced PCA features are extracted and used as the input features to train a PNN classifier to identify devices. The authors collect data from eight different IEEE 802.11b WiFi transmitters at a sampling rate of 5 GSps. A total of 100 transmissions are captured from each of the eight transmitters. To study the effect of sampling rate, authors use the decimation process to downsample the 5 GSps to 2.5 GSps, 1 GSps, 500 MSps, 200 MSps, 100 MSps, 50 MSps, and 28 MSps.
\\
The classification accuracy is evaluated by conducting two experiments. In the first experiment, downsampling was performed after detection of the transient at a higher sampling rate. In the second experiment, transient detection was performed after downsampling the original signal. The average classification accuracy of both amplitude and PCA features at all the sampling rates was $97.7\%$ and $97.5\%$ for the first and second experiment, respectively, indicating that sampling rates have very little to no impact for transient based RF fingerprinting of WiFi transmitters.

\subsection{Attentional Learning}

Attention mechanism was first introduced for RF fingerprinting in \cite{AJagannath22GLOBECOM}. The authors adopt a cross-domain attentional architecture that extracts spatio-temporal, temporal, and time-frequency features from $1024\times1$ raw IQ input samples. A 1D/2D CNN architecture in conjunction with gated recurrent units (GRUs) and STFT processing were adopted to extract the multiple domain features from the raw IQ samples. Further, the capability of the model to perform multiple tasks (emitter and protocol recognition) were demonstrated with the MTL version of the proposed architecture. The authors perform real-world experimental evaluation under single day train-test (TTSD) and mixed days train-test (TTMD) scenarios. Here, the authors consider real-world IoT devices such as Raspberry Pis and Lenovo laptops with combo chipsets (emitting Bluetooth and WiFi waveforms from a single chipset) and achieve a fingerprinting accuracy of 84.3\% and 63.8\% under the TTSD and TTMD scenarios respectively  (while performing 100\% protocol recognition) in identifying emissions from 10 COTS chipsets. 

A recent work in attentional learning for Bluetooth fingerprinting was reported in \cite{AJagannath22Tech}. Here, the authors tuned into 2 MHz of the challenging frequency hopping Bluetooth spectrum for identifying 10 COTS Bluetooth emitters. A scalable, hybrid CNN-GRU architecture with ability to support input tensor length of up to 1 MS is proposed. The authors demonstrate the computational efficiency of the proposed architecture in contrast to the benchmark model and report a $16.9\times$ fewer floating point operations (FLOPs) and $7.5\times$ lesser trainable parameters. The significance of processing greater sample lengths in identifying the challenging frequency hopping Bluetooth waveform was elaborately studied with reported accuracy of up to 91\% in identifying 10 COTS emitters.

\subsection{Open RF Fingerprinting Dataset}
\label{sec:dataset}
A soaring issue in the applied AI/ML for RF realm, unlike other prominent fields such as computer vision and natural language processing, is the lack of availability and uniformity of diverse and large-scale datasets which can be integrated as well as easily ported to AI/ML frameworks such as Keras \cite{kerasdata}, PyTorch \cite{torchdata}, TensorFlow \cite{tfdata}, etc. A few datasets have been released recently to aid deep learning for wireless communications \cite{oshea2018,crawdad,hisarmod,AjagannathComnet2021} for modulation and protocol classification, however, due to the lack of momentum and a common standard to organize the datasets in contrast to computer vision and NLP in AI/ML, these are not integrated yet with such frameworks. Another factor contributing to the dataset inaccessibility is the obliviousness of practitioners in this field of the available datasets, although limited. Accordingly, here we present a summarized excerpt on each of the openly available RF fingerprinting dataset to educate the reader.
\subsubsection{Large-scale Bluetooth dataset from 86 smartphones} In \cite{bt86phones}, the authors present an elaborate database comprising Bluetooth RF recordings from COTs smartphones of different makes and models captured at different sampling rates. The dataset contains recordings captured over the period of several months since the unique fingerprint from hardware impairments do not vary significantly over short time spans - days, weeks, or months. The smartphones were kept at a fixed distance of 30cm from the receiving antenna connected to a high sampling rate oscilloscope (Tektronix TDS7404) along with a low resolution 8-bit ADC. Since the Bluetooth operate at the ISM2400 band, a COTS antenna operating in this frequency range was utilized. The edge detection mode of the oscilloscope was leveraged to record the samples of duration 10$\mu$s into a text format (.txt). The recorded samples are real-valued time series (voltage/times). The entire database is split into sub-datasets comprising Bluetooth signals sampled at 5 GSps, 10 GSps, 20 GSps, and 250 MSps. Corresponding to each dataset,150 Bluetooth signals from each device was recorded yielding a total of 12,900 captures from 86 smartphones. The authors also note that the spur signals introduced by the oscilloscope were removed by band pass filtering. Moreover, the filtered samples are normalized such that the samples are in the range of -1 to +1.

\subsubsection{Dataset containing RF signals from 17 drone remote controllers} The authors of \cite{droneRC17} released a RF signal dataset to enable researchers to develop UAV identification techniques based on the signal captured from the remote controllers. The communication between UAV and the remote controller can enable AI/ML frameworks to effectively fingerprint UAVs. The captures were recorded by placing the drones in an idle state such that only the remote controller data is captured. The receiver frontend comprised a 6 GHz bandwidth Keysight MSOS604A oscilloscope, 2.4 GHz 24 dBi grid parabolic antenna, and a low-noise amplifier operating from 2 GHz to 2.6 GHz. The distance between the drone remote controller and the receiving antenna was varied from 1m to 5m. The RF signal is recorded as digitized voltage vs time samples at a sampling rate of 20 GSps with 5 Million samples per signal. The waveforms comprise emissions from 17 drone remote controllers from eight different manufacturers. The database is containerized in a MATLAB (.mat) format.

\begin{table*}
\centering
\footnotesize
\small
\caption{Summary table of open RF fingerprinting datasets}
\begin{tabular}{m{28pt}m{58pt}m{22pt}m{45pt}m{43pt}m{30pt}m{52pt}m{40pt}}
\hline
Dataset & Waveform & Emitter count & Emitter type & Receiver & Dataset format  & Generated\;/ \newline Real-world & Frequency\\
\hline \hline

\cite{bt86phones} & Bluetooth & 86 & Smartphones & Tektronix\newline TDS7404 & .txt  & Real-world & 2.4\;GHz \\\\
\cite{droneRC17} &Non-standard &17 &Drone\;remote\newline controllers &Keysight\newline\;MSOS604A &.mat  & Real-world & 2.4\;GHz\\\\
\cite{adsb100} & ADS-B & 100 & Commercial aircraft & BladeRF & .mat  & Real-world & 1090\;MHz \\\\
\cite{adsb130} & ADS-B & $>140$ & Commercial aircraft & USRP\;B210 & .mat  & Real-world & 1090\;MHz \\\\
\cite{oracle_dataset} &IEEE\;802.11a &16 & USRP\;X310 & USRP\;B210 & SigMF  &Generated &2.45\;GHz\\\\
\cite{hovering_uavs_dataset} &Non-standard &7 &DJI\;M100 &USRP\;X310 &SigMF  &Generated &2.4065\;GHz\\\\
\cite{powder_dataset} & IEEE\;802.11a,\newline LTE, 5G-NR &4 &USRP\;X310 & USRP\;B210 &SigMF &Generated &2.685\;GHz\\\\
\cite{shawabka_dataset} & IEEE\;802.11a/g &20 &USRP\;X310\newline USRP\;N210 &USRP\;N210 & SigMF  &Generated &2.432\;GHz\\\hline

\end{tabular}
\label{table:RFTable}
\end{table*}

\subsubsection{Real world ADS-B signals dataset from over 140 commercial aircrafts} A real world dataset containing ADS-B signal emissions from more than 140 commercial aircrafts to air traffic control (ATC) centers is provided by the authors of \cite{adsb130,adsb140}. Commercial aircrafts broadcast their geographical coordinates along with their unique  International  Civil  Aviation  Organization  (ICAO) identifiers to the ATC centers using ADS-B standard. The ADS-B signals are captured with a USRP B210 receiver tuned to 1090 MHz and at a 8 MSps sampling rate over a period of 24 hours at the Daytona Beach  international  airport. The authors decoded the ADS-B messages to extract the aircraft identity codes and utilized the messages from over 140 most frequently seen aircrafts to form the dataset. The authors have another dataset of ADS-B waveforms from 100 aircrafts at \cite{adsb100} received with a BladeRF SDR. Both datasets are containerized into MATLAB (.mat) format.

\subsubsection{ORACLE RF Fingerprinting Dataset of IEEE802.11a from 16 emitters}
\label{sec:dataset_ORACLE}
The authors of \cite{oracle_dataset} present a WiFi IEEE 802.11a emitter dataset to detect unique radios using ORACLE RF fingerprinting approach. The dataset contains two sets: Dataset\#1 and Dataset\#2. Dataset\#1 consists of IEEE 802.11a standard Wireless Local Area Network (WLAN) frame IQ samples from 16 USRP X310 SDRs collected using a USRP B210 Radio sampling at a rate of 5 MSps at a center frequency of 2.45 GHz. For each of the 16 transmitters, the IQ samples are captured at a varying transmitter-receiver distance from 2 ft - 62 ft in steps of 6 ft. Dataset\#2 consists of demodulated IQ symbols with intentional impairment introduction such that the synthetic hardware impairments dominate the channel effects. Accordingly, the authors use the GNU Radio function \emph{set\_iq\_balance} to introduce intentional IQ imbalance (16 IQ imbalance configurations corresponding to 16 emitters) to the transmit chain of the RF daughterboard.  The recording are the demodulated IQ symbols after equalizing over-the-cable transmissions from USRP X310s collected using USRP B210 Radio. Both the datasets are formatted according to SigMF specifications wherein each data file in binary format is accompanied by a JSON metadata file.

\subsubsection{Non-standard Waveforms from 7 Hovering Unmanned Aerial Vehicles (UAVs)}
\label{sec:dataset_UAV}
In \cite{hovering_uavs_dataset}, the authors create a dataset for RF fingerprinting of hovering UAVs. The dataset consists of signals collected from 7 identical DJI M100 UAVs in an RF anechoic chamber. Signals are captured using an USRP X310 with UBX 160 USRP daughterboard. The receiver is tuned to the 10 MHz of downlink channel centered at 2.4065 GHz. The signals are captured by flying the UAVs individually at distances of 6, 9, 12, and 15 ft from the receiver. Each capture consists of 4 cycles of recording IQ samples for $\sim2$ seconds and pausing for $\sim10$ seconds, resulting in 4 non-overlapping bursts with $\sim140$ interleaved short periods of data and noise in each burst. Accordingly, with a total of 7 UAVs where each are flew at 4 distances with 4 bursts (each of $\sim$ 140 examples) at each distance, the dataset provides over 13k examples of $\sim$ 92k IQ samples per example. The dataset is in SigMF format with data of each capture stored in binary format accompanied by a JSON file with the metadata of the capture.

\subsubsection{RF Fingerprinting on the POWDER Platform with 4 emitters}
\label{sec:dataset_POWDER}
The authors of \cite{powder_dataset} provide a dataset from 4 different emitters transmitting waveforms belonging to 3 wireless standards to demonstrate and evaluate feasibility of RF fingerprinting of base stations with a large-scale over-the-air experimental POWDER platform \cite{powder_platform}. Using a fixed endpoint (Humanties) USRP B210 receiver, IQ samples are collected from four emitters in the POWDER Platform: MEB, Browning, Beavioral, and Honors. The emitters are bit-similar USRP X310 radios which transmit standard compliant IEEE 802.11a, Long Term Evolution (LTE), or 5G New Radio (5G-NR) frames generated using WLAN, LTE, and 5G toolboxes from MATLAB, respectively. The USRP B210 receiver is tuned to record 2.685 GHz (Band 7) at a sampling rate of 5 MSps for WiFi and 7.69 MSps for LTE and 5G. On two independent days, five sets of 2 s of IQ samples are recorded from each of the links. Consequently, the dataset is organized into a Day-1 and Day-2 sets. The dataset follows the SigMF specifications.
 
\subsubsection{Exposing the Fingerprint Dataset}
\label{sec:dataset_exposing_fingerprint}
Al-Shawabka et al. create and share a dataset\cite{shawabka_dataset} for experimenting and evaluating radio fingerprinting algorithms. WiFi standard IEEE 802.11a/g signals are collected from 20 National Instruments SDR (12 NI N210 and 8 NI X310) running GNU Radio. Four datasets are created with three different channel conditions and two different environments. Dataset "Setup 1" consists of signals captured from 20 transmitting SDRs with each transmitter using a dedicated Ettus VERT2450 antenna and varying distance from the receiver. The dataset collection process is repeated on ten days. Dataset "Setup 2" is captured similarly to "Setup 1", but all the SDR use a common Ettus VERT2450 antenna making all 20 devices equidistant from the receiver. This leads to all transmissions experiencing similar channel and multi-path conditions. The dataset collection process is repeated on two different days. Dataset "Setup 3" is collected by capturing the WiFi signals from 20 transmitters using a single coaxial RF SMA cable and a 5 dB attenuator. Thereby, all signals experience the same channel conditions and eliminate any multipath conditions. The dataset collection process is repeated on two different days. Datasets "Setup 1", "Setup 2", and "Setup 3" are collected in an arena "in the wild" environment. Dataset "Setup 4" is collected similar to "Setup 2" but in an anechoic chamber with each transmitter connected to the same antenna. All the "Setup 4" IQ samples are collected on one day. The following three IQ samples are collected: Raw IQ before FFT, Raw IQ after FFT, and Equalized IQ for all the datasets. Each of the IQ sample files is labeled using SigMF and is accompanied by a JSON file containing the metadata of each of the transmission settings.

A tabular summary of the openly available RF fingerprinting datasets is presented in Table \ref{table:RFTable} to allow the reader to contrast the distinguishing features. Although the datasets \cite{oracle_dataset,powder_dataset,hovering_uavs_dataset,shawabka_dataset} are synthetically generated, they follow the SigMF specifications allowing easy integration into AI/ML frameworks in contrast to the other discussed real-world datasets which would require specific import scripts requiring MATLAB or csv readers.

\section{Research Challenges and Future Direction}
\label{sec:open}
\begin{figure}[h]
\centering
\includegraphics[width=0.99\columnwidth]{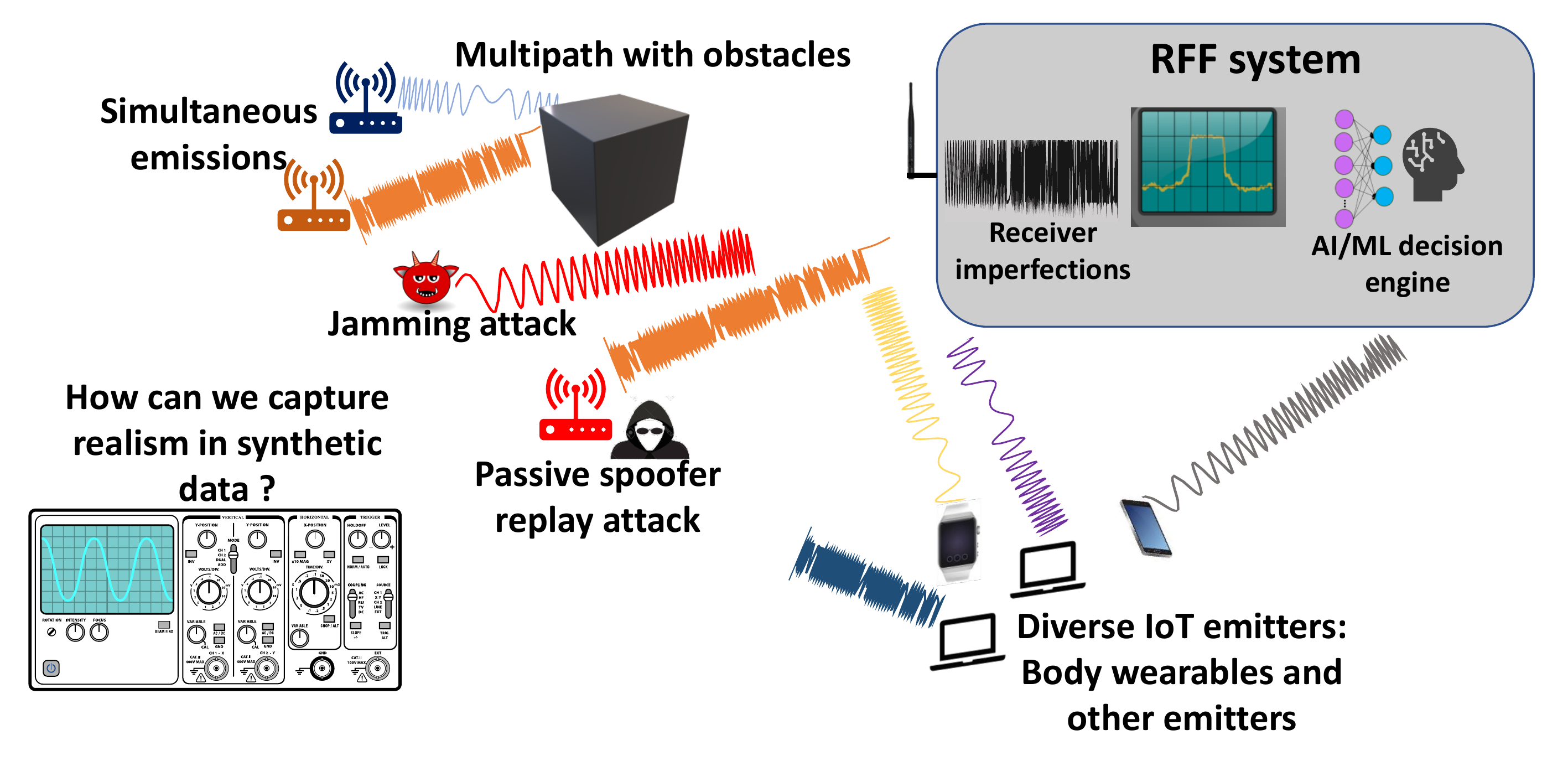} 
\caption{Wireless IoT fingerprinting challenges}
\label{fig:iot}
\end{figure}

In so far, we have seen the various wireless device fingerprinting approaches and how it plays a role in wireless security. For completeness of the presented subject, in this section, we motivate future research in this direction by identifying a few key open research problems and opportunities towards developing a robust radio frequency fingerprinting system (RFFS). These challenges are also illustrated in an IoT network setting in Figure \ref{fig:iot} to ease the reader into the potential research avenues.

\emph{\textbf{Impact of receiver hardware:}} Similar to how the transmitter hardware introduce unique distortions, the receiver hardware that captures and processes these emissions for fingerprinting can impact the fingerprinting approach. 
Specifically, the phase noise, clock offsets, filter distortions, IQ imbalance, etc., introduced by the receiver hardware could etch its own unique fingerprint to transform the transmitter fingerprint to appear as from a rogue or unidentified emitter. The ADC sampling rate as well as bandwidth of low pass filter (LPF) play an equally important role in retaining the fingerprint features that reside in the side lobes of the power spectrum density (PSD). Higher sampling rates were shown to retain the fingerprint features at a cost of increased noise using actual MicaZ sensors \cite{wangUB}. Moreover, the effect of antenna polarization and orientation at the transmitter and receiver end can cause fluctuations in radiation pattern affecting the fingerprint extraction performance. The imperfection of emitter antenna hardware can also contribute to the fingerprint feature set enabling wireless emitter identification \cite{danev2009physical}. We argue here that the number of receiver antennas, type, their orientation, and polarization can impact the classification performance of the fingerprinting system.

One way to tackle this in a supervised learning setting would be to incorporate captures from multiple receiver hardwares corresponding to an emitter in the dataset. Such a larger distribution of training data would allow the model to generalize and differentiate the emitter fingerprint from the recorded waveforms. The independence of fingerprinting algorithm can be assessed by training on samples captured by a particular receiver hardware and evaluating the learned emitter features by testing on samples from another receiver hardware.

\emph{\textbf{Vulnerabilities of RFFS}}:  
The \textit{broadcast} nature of wireless emitters renders them exposed and susceptible to identity spoofing. Few such attacks are DoS, impersonation, bandwidth theft, etc. It is often overlooked that passive receiver threats can build up their own dataset of emissions from specific transmitters to build cognitive RFFS. Developing or perturbing the emitter fingerprint such that it cannot be extracted by passive listeners while allowing only legit receivers to extract or identify the signature is another interesting research problem to enhance wireless security. Generally, it is assumed that RF fingerprinting is robust to impersonation attacks due to the difficulty in reproducing the frontend impairments with replay attacks since that will introduce the hardware defects of the replaying device. This area is pristine and the research here is limited currently. In literature, it was shown that transient-based RF fingerprinting is more resilient to impersonation attacks in contrast to modulation-based RF fingerprinting \cite{phyattacks_1}. Another work in \cite{phyattacks_2} analyzed the effect of several low-end receivers (manufactured with inexpensive analog components) on the resilience of modulation-based RF fingerprinting to impersonation attacks. Their evaluation revealed that the RF fingerprint from a specific transmitter varies across the receivers. The receiver's hardware imperfections as we have discussed above contributes to the fingerprint feature set. Further, they exploit this fact to thwart the impersonation attacks and state that the impersonator would not be able to extract the fingerprint features contributed by the receiver hardware, rendering an even robust RFFS. Another threat that can disrupt the RFFS are jamming DoS attacks whereby the intruder can continuously transmit in the operating frequency. This area will require more analysis to evaluate the resiliency of RFFS to such DoS attacks.

On the flip side, jamming can also be used as a defense strategy to mask the RF fingerprint of transmitters for covert and confidential operations. \textit{RF fingerprint obfuscation} - such that the fingerprint can only be extracted by the legitimate receiver while remaining undetectable to others - was experimentally studied on WiFi signals in \cite{rf-veil}. The authors achieve this by introducing randomized phase errors such that only the legitimate receivers with a preshared key and randomization index can decode the message as well as the fingerprint.

\emph{\textbf{Robustness in realistic operation environment}} The fingerprinting literature to date (at the time of writing this article) has only looked at the problem of identifying emitter signature when only one emitter is active. An even challenging problem would be when multiple emitters are active, this is typical of a real-world setting. Such a scenario would require the fingerprinting algorithm to separately distinguish and extract the signatures of each emitters from the received signal clutter. Another challenge involved in studying such a scenario would be the availability of a dataset that incorporates multiple active emitters. Each emitter transmission creates its own propagation path from the transmitting antenna to the radio frontend of the receiver hardware. The effect of multipath propagation effects and location of the emitter relative to the receiver is enough to create a unique signature which would vary with location and wireless channel effects. These dynamic fading and location effects due to its inherent randomness could mask the \emph{pure} emitter signature leading to false alarms and misclassification. 

In \cite{wangUB} authors demonstrate the effect of small scale and large scale fading on the PSD. It was illustrated that the side lobes of the PSDs that carry the most identity information were significantly distorted due to multipath channel effects when the sensors were far apart than when they were in close proximity to the receiver. Equalization at the receiver that would compensate for the multipath effects without deteriorating the fingerprint features is still an open research problem. 

\emph{\textbf{Simulation-reality gap}}: An equally important point  to consider is the realism in the generated or synthetic data. The generalization of deep learning models to actual radio emissions after being trained on synthetic data is difficult to achieve. Such a capability gap arises from the assumptions in terms of the transmitter hardware imperfections and fading channel while generating the synthetic dataset in contrast to the actual hardware and environmental effects. 

A towering issue that leads to generating synthetic data is the lack of or limited access to real-world data from actual IoT sensors and radios. This is not the case in more popular machine learning fields such as natural language processing (NLP) and computer vision (CV) where a plethora of large-scale datasets such as MNIST \cite{mnist}, Stanford sentiment \cite{stanford}, IMDb \cite{imdb}, Sentiment140 \cite{sentiment}, etc., are readily available. As highlighted in section \ref{sec:dataset}, there are several recent efforts to mitigate this challenge. Further, the lack of a uniform standard for the dataset structure and organization stymies the adoption of existing datasets to different machine learning frameworks. We state here that training the neural networks with a larger distribution of data is the key to a generalized performance. Generalization is the first step to deployment-ready fingerprinting solutions.

\section{Conclusion}
This article presented a systematic review of the RF fingerprinting approaches over the past two decades by first broadly classifying them into traditional and DL-based followed by dissecting each in a categorized manner. We first provided context to the reader by introducing and summarizing the three pillars of SIGINT - modulation recognition, protocol classification, and emitter identification. We present an invaluable and concise discussion on the diverse applications of RF fingerprinting to highlight the practical use cases of the subject under study. To elucidate and dilute the vast literature on traditional and DL-based fingerprinting approaches, we present a categorized and clear layout of each. We have also provided tabular comparative study of the reviewed works wherever applicable for summarizing in a straightforward manner. In order to equip the reader with the essential toolkit to delve into this topic, we reviewed the most relevant DL approaches in a tutorial manner prior to diving into the DL-based fingerprinting techniques. Since the knowledge of and access to openly available datasets are key to practice the reviewed approaches, we have provided an elaborate discussion on the most relevant RF fingerprinting datasets. Finally, in order to stimulate future research in this realm, we present a roadmap of potential research avenues in an illustrative manner.


%





\ifCLASSOPTIONcaptionsoff
  \newpage
\fi



%



\bibliographystyle{IEEEtran}

\bibliography{Survey}

%



\begin{IEEEbiography}[{\includegraphics[width=1in,height=1.25in,clip,keepaspectratio]{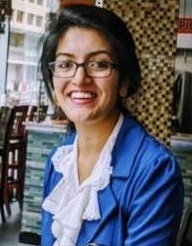}}]{Anu Jagannath} currently serves as the Founding Associate Director of Marconi-Rosenblatt AI/ML Innovation Lab at ANDRO Computational Solutions, LLC. She received her MS degree from State University of New York at Buffalo in Electrical Engineering. She is also a part-time PhD candidate and is with the Institute for  the Wireless Internet of Things at Northeastern University, USA. Her research focuses on MIMO communications, deep machine learning, reinforcement learning, adaptive signal processing, software defined radios, spectrum sensing, adaptive physical layer, and cross layer techniques, medium access control and routing protocols, underwater wireless sensor networks, and signal intelligence. She has rendered her reviewing service for several leading IEEE conferences and Journals. She is an IEEE Senior Member. She is the co-Principal Investigator (co-PI) and Technical Lead in multiple Rapid Innovation Fund (RIF) and SBIR/STTR efforts involving  applied AI/ML for wireless communications. She is also the inventor on 6 US Patents (granted and pending). 
\end{IEEEbiography}

\begin{IEEEbiography}[{\includegraphics[width=1in,height=1.25in,clip,keepaspectratio]{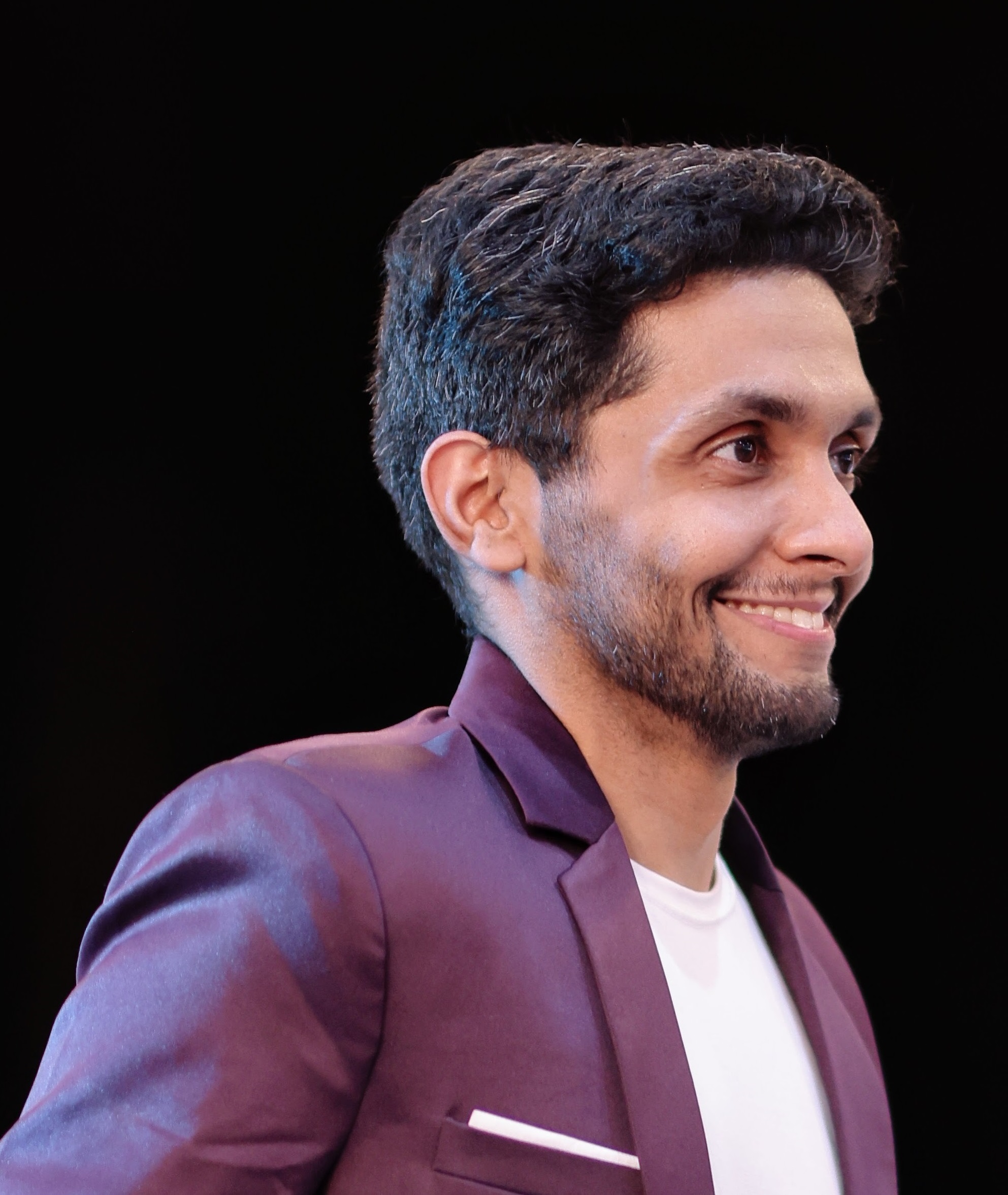}}]{Jithin Jagannath} is the Chief Scientist of Technology and Founding Director of the Marconi-Rosenblatt AI/ML Innovation Lab at ANDRO Computational Solutions. He is also the Adjunct Assistant Professor in the Department of Electrical Engineering at the University at Buffalo, State University of New York. Dr. Jagannath received his B. Tech in Electronics and Communication from Kerala University;  M.S. degree in Electrical Engineering from University at Buffalo, The State University of New York; and received his Ph.D. degree in Electrical Engineering from Northeastern University. Dr. Jagannath was the recipient of 2021 IEEE Region 1 Technological Innovation Award with the citation, "For innovative contributions in machine learning techniques for the wireless domain''. He is also the recipient of AFCEA International Meritorious Rising Star Award for achievement in engineering and AFCEA 40 Under 40 award. 

Dr. Jagannath heads several of the ANDRO's research and development projects in the field of Beyond 5G, signal processing, RF signal intelligence, cognitive radio, cross-layer ad-hoc networks, Internet-of-Things, AI-enabled wireless, and machine learning. He has been the lead and Principal Investigator (PI) of several multi-million dollar research projects. This includes a Rapid Innovation Fund (RIF) and several Small Business Innovation Research (SBIR)s for several customers including the U.S. Army, U.S Navy, Department of Homeland Security (DHS),  United States Special Operations Command (SOCOM). He is currently leading several teams developing commercial products such as SPEARLink\texttrademark, DEEPSPEC\texttrademark~ among others. He is an IEEE Senior Member and serves on the IEEE Signal Processing Society's Applied Signal Processing Systems Technical Committee. Dr. Jagannath's recent research has led to several peer-reviewed journal and conference publications. He is the inventor of 12 U.S. Patents (granted and pending). He has been invited to give various talks including Keynote on the topic of machine learning and Beyond 5G wireless communication. He has been invited to serve on the Technical Program Committee for several leading technical conferences.
\end{IEEEbiography}


\begin{IEEEbiography}[{\includegraphics[width=1in,height=1.25in,clip,keepaspectratio]{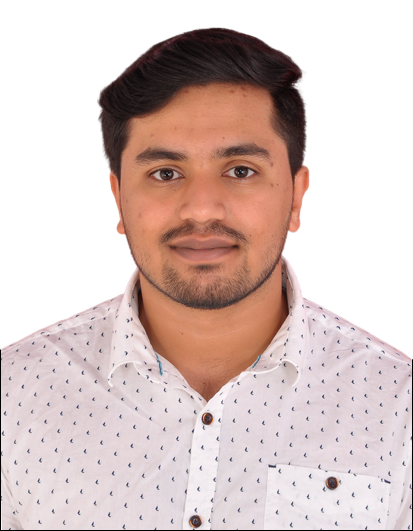}}]{Prem Sagar Pattanshetty Vasanth Kumar}
is an Associate Scientist/Engineer of Marconi-Rosenblatt AI/ML Innovation Lab at ANDRO Computational Solutions, LLC. He received his Bachelor of Engineering degree in Electronics and Communication from Visvesvaraya Technological University and a Master of Science degree in Electrical Engineering from State University of New York at Buffalo. His areas of research interests include machine learning,  reinforcement learning, neural networks, software defined radios, physical layer, and wireless communications.
\end{IEEEbiography}




\end{document}